\documentclass[Afour,sageh,times]{sagej}

\usepackage{moreverb,url}

\usepackage[colorlinks,bookmarksopen,bookmarksnumbered,citecolor=red,urlcolor=red]{hyperref}

\usepackage{algorithm2e}

\usepackage{graphicx}
\usepackage{caption}
\usepackage{subcaption}
\usepackage{array}
\usepackage{multirow}
\usepackage{xcolor}

\newcommand\BibTeX{{\rmfamily B\kern-.05em \textsc{i\kern-.025em b}\kern-.08em
T\kern-.1667em\lower.7ex\hbox{E}\kern-.125emX}}

\def\volumeyear{2023}

\usepackage{enumitem}

\usepackage[toc,page]{appendix}

\begin{document}

\runninghead{Chaffre et al.}

\title{Sim-to-Real Transfer of Adaptive Control Parameters for AUV Stabilization under Current Disturbance
}

\author{Thomas Chaffre\affilnum{1} Jonathan Wheare\affilnum{1} Andrew Lammas\affilnum{1} Paulo Santos\affilnum{1,3}\\Gilles Le Chenadec\affilnum{2} Karl Sammut\affilnum{1,3} and Benoit Clement\affilnum{1,2,3}}

\affiliation{\affilnum{1}Flinders University, College of Science and Engineering, Adelaide, Australia\\
\affilnum{2}ENSTA Bretagne, Lab-STICC UMR CNRS 6285, Brest, France\\
\affilnum{3}CROSSING IRL CNRS 2010, Adelaide, Australia}

\corrauth{Thomas Chaffre, Flinders University, Adelaide, Australia.}

\email{thomas.chaffre@flinders.edu.au}

\begin{abstract}
Learning-based adaptive control methods hold the premise of enabling autonomous agents to reduce the effect of process variations with minimal human intervention. However, its application to autonomous underwater vehicles (AUVs) has so far been restricted due to 1) unknown dynamics under the form of sea current disturbance that we can not model properly nor measure due to limited sensor capability and 2) the nonlinearity of AUVs tasks where the controller response at some operating points must be overly conservative in order to satisfy the specification at other operating points. Deep Reinforcement Learning (DRL) can alleviates these limitations by training general-purpose neural network policies, but applications of DRL algorithms to AUVs have been restricted to simulated environments, due to their inherent high sample complexity and distribution shift problem. This paper presents a novel approach, merging the Maximum Entropy Deep Reinforcement Learning framework with a classic model-based control architecture, to formulate an adaptive controller. Within this framework, we introduce a Sim-to-Real transfer strategy comprising the following components: a bio-inspired experience replay mechanism, an enhanced domain randomisation technique, and an evaluation protocol executed on a physical platform. Our experimental assessments demonstrate that this method effectively learns proficient policies from suboptimal simulated models of the AUV, resulting in control performance 3 times higher when transferred to a real-world vehicle, compared to its model-based nonadaptive but optimal counterpart.

\end{abstract}


\maketitle

\section{Introduction}

Recently there has been a growing presence of autonomous vehicles in various sectors of society \citep{Hakak2022AutonomousVI,DBLP:journals/corr/abs-2301-01755,Wibisono2023ASO}. Whether it is cars, trains, warehouse robots, or delivery quadcopters, the field of autonomous vehicles is flourishing. This progress is driven by the desire to enhance productivity, accuracy, and operational efficiency, while also prioritising the safety of human operators and users. Although this trend is observed in various domains, there is a noticeable discrepancy in the development of underwater applications. Despite similar requirements for tasks such as offshore platform inspections, marine geoscience, coastal surveillance, and underwater mine countermeasures, most unmanned underwater vehicles still rely on remote operation or possess limited autonomy capabilities. This issue is even more pronounced in the context of small-sized autonomous underwater vehicles (AUVs). These vehicles are required to operate over large regions (from deep oceans to coastal and riverine regions), and over lengthy periods of time (extending from several hours to days before the possibility of human intervention) performing complex tasks such as search and rescue \citep{AC05}, underwater manipulation \citep{Marani2009UnderwaterAM}, pipeline and facility inspection operations \citep{Gilmour2012FieldRA}, target following \citep{Sun2015TargetFF}, under-ice exploration \citep{Barker2020ScientificCA}, among others. Nevertheless, the autonomous control of underwater autonomous vehicles still presents several challenges. One of which is the limited range and bandwidth of underwater communication systems, which can hinder real-time control and data transmission. Additionally, the underwater environment is harsh, with factors such as high pressure, corrosive saltwater, and poor visibility, which can affect the performance and reliability of control systems. Navigating accurately and avoiding obstacles in underwater environments is also more complex due to the presence of current disturbances. Developing robust control algorithms that can handle these challenges and ensure the safe and efficient operation of underwater autonomous vehicles remains an ongoing area of research and development. 

The present work assumes the standpoint of learning-based adaptive control methods, where machine learning algorithms are employed to compensate for the unknown aspects of a process while control over the known parts is ensured by using traditional methods. This research focuses on the control of manoeuvring tasks for autonomous underwater vehicles (AUVs), specifically the stabilisation of the vehicle at a fixed velocity and orientation. The AUV is assumed to be fully actuated and affected by external disturbances, represented by sea currents, which are considered non-observable variables in this research. The dynamics of an AUV can be described as a combination of its known component and the unknown component. To address this, the present paper builds upon our previous work \citep{CHAFFRE2022}, whereby a novel deep reinforcement learning method was used to model the unknown part of the plant, whereas a traditional PID controller was used to model its known part.

Reinforcement learning (RL) \citep{Sutton-Barto2ed}, a subfield of machine learning (ML), focuses on developing algorithms and techniques for an agent to learn how to make sequential decisions in an environment to maximise a cumulative reward. It draws inspiration from behavioural psychology, where the agent learns through trial and error by interacting with its environment and receiving feedback in the form of (positive or negative) rewards. When applied to real robots, a major challenge in reinforcement learning is to successfully transfer policies, learned from simulated environments, to the target domain. Although there have been significant advances in the development of Deep Reinforcement Learning (DRL), which extends RL by combining RL algorithms with deep neural networks improving the scalability and generalisation of methods, sim-to-real transfer remains a bottleneck \citep{zhao00}.

\begin{figure}[htbp]
\centering
\captionsetup{justification=justified}
\begin{subfigure}{0.4\textwidth}
\centering
\includegraphics[width=\textwidth]{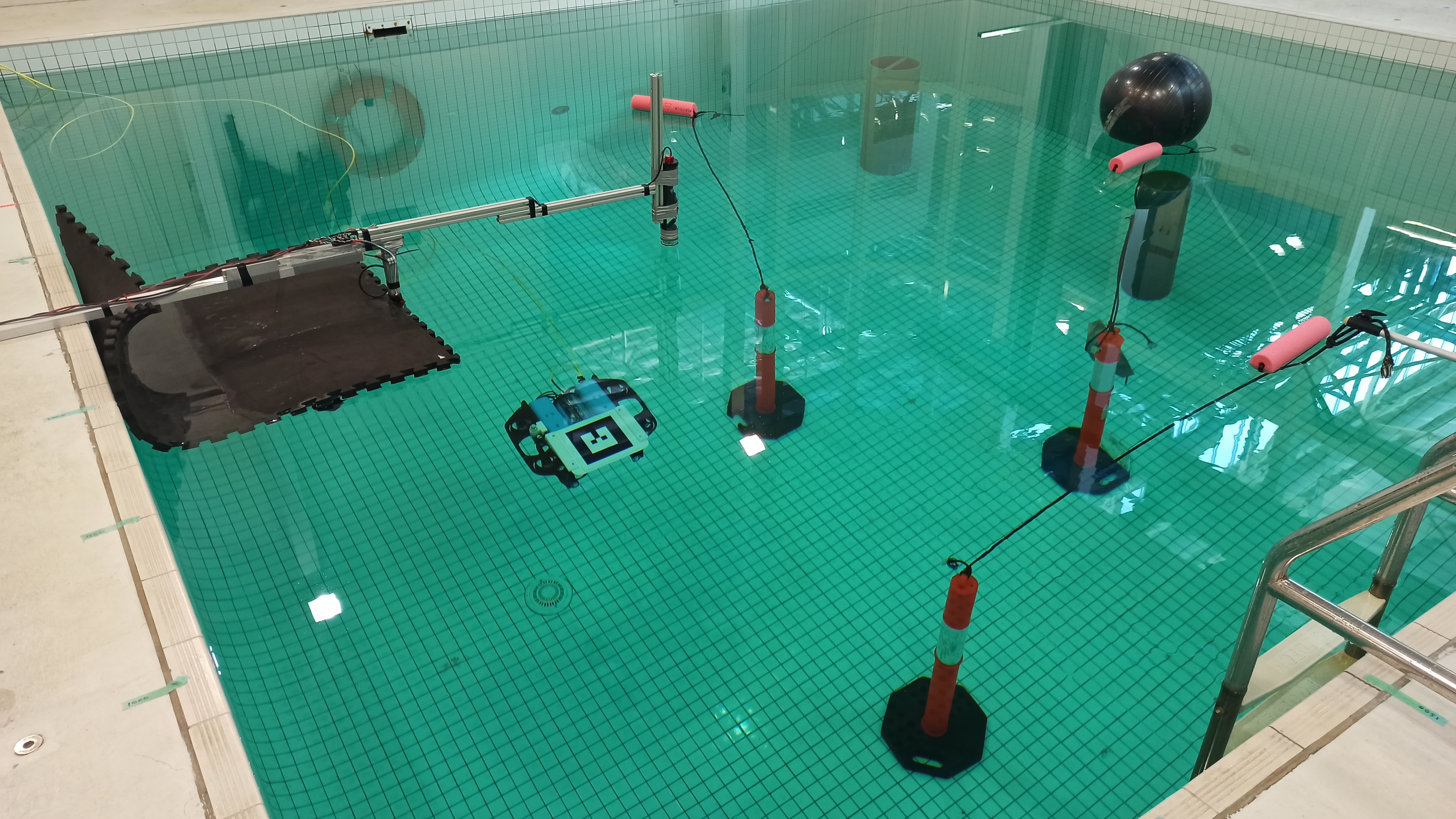} 
\caption{The camera is placed facing down and at the same position using locks that are engraved on the room floor.}
\label{cam-1}
\end{subfigure}\\
\vspace*{0.25cm}
\begin{subfigure}{0.4\textwidth}
\centering
\includegraphics[width=\textwidth]{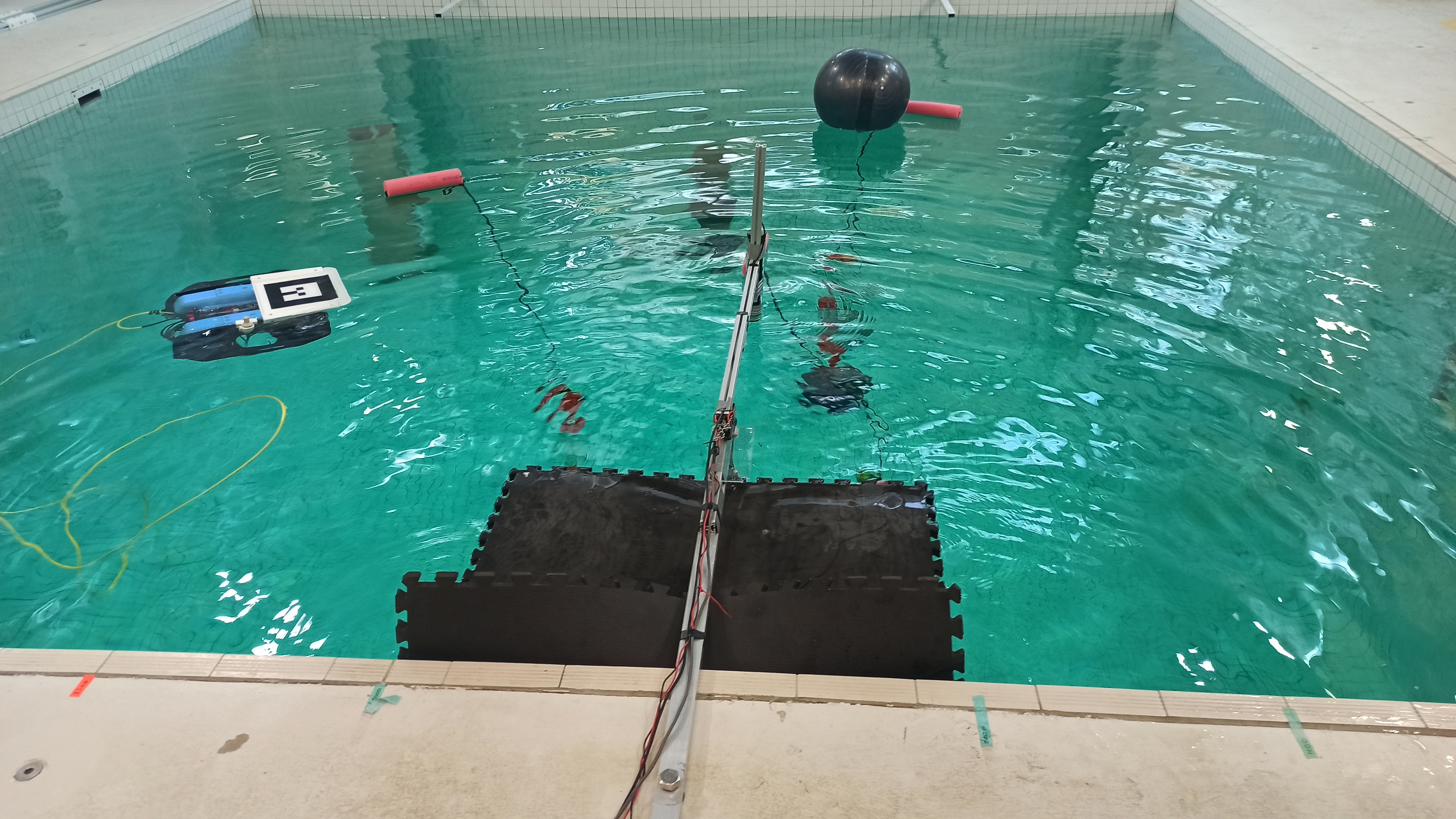}
\caption{The two thrusters are strong enough to produce water displacement all over the tank.}
\label{exp-env}
\end{subfigure}\\

\caption{Illustration of the setup for the experiments. We collected over 180,000 timesteps from the experiment to evaluate the predictive model.}
\label{experiments-settings-1}
\end{figure}

The focus of the present paper is on the experimental evaluation of transferring policies that were first learned by a DRL agent in a virtual environment to a physical AUV under various disturbance regimes. The DRL algorithm used in this work is the Soft Actor-Critic with Automatic Temperature Adjustment algorithm \citep{haarnojaSoft3}, which was combined in this work with the Biologically-Inspired Experience Replay (BIER) method introduced in \citep{CHAFFRE2022}. BIER is a Replay Mechanism \citep{Lin2004SelfimprovingRA} that incorporates two distinct memory buffers: one that stores and replays incomplete trajectories of state-action pairs and another that prioritises high-quality regions of the reward distribution. BIER was used to find the control parameters for a virtual AUV, whose policies were transferred to a real vehicle (a BlueRov AUV \citep{bluerov2}) in an indoor pool environment with two thrusters specifically used here to generate current disturbance in the tank. This experimental environment is depicted in Figure \ref{exp-env}. 

In this context, the task of underwater multi-station keeping by the AUV was considered in this work.
The summary of our contributions is as follows:
\begin{enumerate}[label=(\Alph*)]
\item Evaluating on a real platform a learning-based adaptive controller and its non-adaptive (but optimal\footnote{with respect to Robust Optimal Control Theory \citep{Doyle1995RobustAO}.}) model-based counterpart, and finding that despite being based on the exact same controller structure, the learning-based method provides notable gains.
\item Discovering that despite being trained on a different vehicle model, the learning-based adaptive controller is still able to regulate the AUV when transferred to the real platform.
\item Investigating the connection between the complexity of the source and target domains and finding that randomised environment complexity mitigates the policy variance, which partially explains the improved sim-to-real transfer.
\end{enumerate}
Earlier works investigating the use of DRL for adaptive control of AUVs have often focused on the design of purely model-free adaptive controllers. We begin by identifying the related work in this area along with the challenges associated with the underwater context in the next section.




\section{Related work}


Real-world systems are often characterised by non-linear dynamics and uncertainty in their motion equations, parameters, and system measurements. Learning-based adaptive controllers offer a promising approach to address this challenge by leveraging model-free learning algorithms to approximate the unknown parts of the system model ($f_2$),  tuning control parameters for desired behaviour \citep{lb-book}. Among the various learning techniques, a prominent candidate for that end is reinforcement learning (RL) \citep{Sutton-Barto2ed}. 

RL formulates the control problem as a Markov Decision Process (MDP), represented by the tuple $\langle S, A, T, R \rangle$, where $S$ denotes the set of possible states, $A$ represents the set of actions that the agent can execute, $T$ is the transition function defining the probabilities of reaching successor states, and $R$ represents the reward function \citep{Sutton-Barto2ed}. The RL process can be summarised as follows: first, at each timestep $t$, the agent selects an action $a \in A$ based on the current state $s \in S$. Then, executing the selected action leads to a transition to a new state $s_{t+1} \in S$, and the agent receives a scalar reward $r_t$ that quantifies the quality of the action outcome with respect to a reward function $R(s)$. The goal of RL is to maximise the expected future rewards that the agent could get at each state. The agent updates the value of the selected action based on the received reward, thereby refining the learned policy.

Deep reinforcement learning (DRL) extends the RL methods by utilising deep neural networks (DNNs) to approximate the functions defining the agent's policy and state-action values, enabling the algorithm to handle high-dimensional state spaces and intricate decision-making processes. This marriage of RL and deep learning has led to remarkable breakthroughs, allowing DRL algorithms to excel in a wide range of domains with increasing levels of complexity.

\begin{figure}[htbp]
\centering
\includegraphics[width=0.75\linewidth]{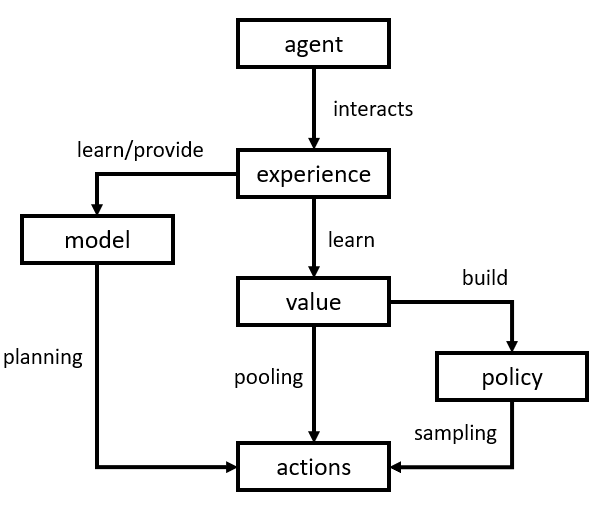}
\caption{Illustration of DRL methods based on the nature of the decision-making process: in model-based methods, the actions are the result of deterministic planning, in value-based methods the actions are pooled over the entire set of possible actions, and in policy gradient methods the actions are sampled from a probability density function.}
\label{RL-methods}
\end{figure}

\textcolor{black}{As illustrated in Figure \ref{RL-methods}, DRL methods commonly consist of some building blocks, and solution methods in DRL are different choices of using them. Some of the most common pieces that people use to put the solution method together are whether or not that solution has:
\begin{itemize}
    \item a \textit{model}, that is something that is explicitly trying to predict what will happen in the environment?
    \item a \textit{value} function, that means is it trying to predict, explicitly, how much reward it will get in the future?
    \item a representation of a \textit{policy}, that means something that is deciding how to pick actions? Is the decision-making process explicitly represented?
\end{itemize}}
\textcolor{black}{The two first classes of solutions methods aim to compute (with model-based methods) or learn (in the case of value-based methods) the Q-value function and then select actions accordingly. Instead, the last class of solutions methods, named Policy Gradient (PG), aims to model and optimize the policy behaviour $\pi$ directly (i.e. the decision-making). The policy is traditionally \citep{Sutton1999PolicyGM} represented by a parameterized function with respect to $\theta$, $\pi_\theta(a|s)$. The value of the reward function $J(\theta)$ depends on this policy and thus we can use various algorithms to optimize $\theta$ which achieves the best reward. The reward function is defined as the expected return and the parameters $\theta$ are optimized with the goal of maximizing the reward function. Modern PG methods use DNNs to approximate the aforementioned functions and are thus denoted as Deep Policy Gradient methods (DPG) and the policy is then denoted as Policy network. DPG methods are currently leading the field of robotics control systems because \citep{Sutton-Barto2ed}:
\begin{itemize}
    \item they have better convergence abilities,
    \item they are effective in high dimensional or continuous action space,
    \item and they can learn stochastic policies.
\end{itemize}}
\textcolor{black}{Below, we present the challenges of using DPG methods in the real world followed by related applications to AUV processes.}

In a recent survey \citep{DulacArnold2020AnEI}, the main challenges of applying DRL to physical robotic systems were listed, including the need for satisfying non-trivial environmental constraints, the high-dimensional (continuous) state and action spaces and the search for efficient solutions to multi-objective reward functions, when dealing with complex problems. The need to satisfy environmental constraints comes from the fact that robots often operate in environments where safety, efficiency, and regulatory requirements must be met. To ensure that the robot's actions comply with these constraints, it is necessary to incorporate them within the DPG framework. Techniques such as constrained (or safe) reinforcement learning \citep{SafeRL}, or integrating constraint satisfaction into the reward function \citep{Spieker2021}, have been employed to address this challenge effectively. Another significant aspect of robotic tasks is the presence of high-dimensional continuous state and action spaces.  Traditional DRL algorithms struggle with these high-dimensional spaces due to the {\em curse of dimensionality} that dictates that, as the number of dimensions or features in an environment increases linearly, the computational and sample complexity of learning algorithms grows exponentially. This results in a dramatic increase in the amount of data required to explore and understand the state space adequately \citep{Berch98}. However, advances in DPG have led to the development of specialised algorithms, such as deep deterministic policy gradients (DDPG) \citep{lillicrap2019continuous}, or  twin-delayed DDPG (TD3) \citep{dankwa20}, based on the Actor-Critic method \citep{Haarnoja2018SoftAA}, that are able to handle high-dimensional continuous spaces more efficiently. Robotic tasks also involve the optimisation of multiple objectives simultaneously. For instance, a robot may aim to complete a task quickly, minimise its energy consumption, and avoid collisions all at once. In order to design reward functions to address all these objectives, Multi-Objective Reinforcement Learning (MORL) \citep{Liu15} has to be considered. In addition, Lyapunov stability with respect to DRL-based control systems is still not fully understood  \citep{Garcia2015ACS}. In fact, in our previous work \citep{HectorKohler}, we compared the Lyapunov stability of a learning-based adaptive controller with a purely adaptive but Lyapunov-based controller in the case of AUV control. We observed that both controllers displayed similar stability in terms of vehicle state stability, with respect to the Lyapunov theory. However, we observed that there is a big contrast in terms of the stability of the controller parameters.
Another challenge arises from the partial observability and non-stationarity of real-world environments. In the context of AUVs in particular, we have a limited capability for onboard sensors due to their small size. This makes it difficult or impossible to measure process disturbances, making it even more challenging to reject them.

Meeting these challenges has been the focus of much work in this area recently. The DDPG \citep{lillicrap2019continuous} algorithm was used to learn the optimal trajectory tracking control of AUVs, where the control problem consists of keeping the error $e=x-x_d$ between the actual trajectory $x$ and the target $x_d$ at zero \citep{8028138}. A loss function was defined to update the parameters of the actor network which includes Lyapunov stability components \citep{Bacciotti2001LiapunovFA}. This approach was compared to a fixed gains PID and the results indicate that the learning-based controller exhibited better performance in terms of tracking error. However, as the stability components were incorporated by an additional term in the actor loss function, there were no formal guarantees that the system would remain stable at all times. Learning-based adaptive control was investigated in \cite{Knudsen2019DeepLF} in a station-keeping task executed by an AUV under unknown current disturbances. The DDPG algorithm was used to control the position of a BlueROV2 platform in surge $x$ and sway $y$ combined to a PD control law that regulated the AUV position in heave $z$ and orientation in roll $\phi$, pitch $\theta$ and yaw $\psi$. The DDPG algorithm was used to learn a PD control law as a function of the vehicle position and velocity at previous timesteps. The training was performed within the  Robot Operating System (ROS) Gazebo simulator \citep{Quigley2009}. The evaluation was conducted on a real platform in an indoor water tank and consisted of three scenarios: two of which had different desired pose definitions and the third assumed a 4-corner test. The first scenario consisted of changing one error state while in the second scenario, both error states were changed at the same time. The 4-corner test consisted of performing station keeping at the 4 corners of a rectangular trajectory. These experiments showed that the agent was able to complete the task under real conditions. The performance was, however, slightly worse in the real environment compared to the simulated one, especially for the most challenging task of a 4-corner test. More recently, Deep Imitation Learning (DIL) \citep{Liu2018ImitationFO,deepmimic} and another Deep Policy Gradient algorithm named Twin Delay Deep Deterministic Policy Gradient (known as TD3) \citep{Fujimoto2018AddressingFA}, were combined in \cite{Chu2020MotionCO} for the design of a learning-based controller for an AUV (the combination of DIL and DRL is denoted as DIRL). The idea of DIL is to use some expert agent to generate examples of appropriate behaviours that are then used to perform the pre-training procedure of the DNNs (in a supervised fashion). Then, the neural networks could be fine-tuned using the normal DRL framework under a reduced number of episodes. They compared the proposed method named IL-TD3 (which stands for the combination of DIL and TD3) under simulation to the original PID controller with and without current disturbances. Results showed that, in the case of no disturbances, both methods were able to solve the task (IL-TD3 exhibited faster response and lower overshoot but at the cost of a much higher thruster solicitation than the PID algorithm). The advantage of their method was also demonstrated with real-life tank experiments on the BlueROV2 platform. When the Policy network was transferred in the real world, which was only trained under simulation, it displayed better performance in depth trajectory following compared to a fixed optimal PID controller.

We can observe that two trends are dominating the field of adaptive control of AUVs, which could be classified into direct and indirect approaches. In the former case, the parameters of the controller are adjusted directly using DRL, whereas in the latter the adjusted control parameters are the result of solving an optimisation problem where the state and/or unknown parameters of the process are first estimated and then used to compute the associated optimal parameters. In most cases, these approaches are applied to classical model-based control structures such as the PD or PID control laws. The objective is then to adjust the parameters of these control structures (their gains) according to process variation, using DRL algorithms such as TD3 and the DDPG algorithms. These deep policy gradient methods build deterministic actors and do not take into account the entropy term from the maximum entropy reinforcement learning framework \citep{Ahmed2019UnderstandingTI}. Most of these works use the original experience replay mechanism \citep{Lin2004SelfimprovingRA}, with a few exceptions, such as \citep{Wang2018ReinforcementLA} where the past experiences of the agent are selected based on different control constraints and stored in separate replay buffers. By using only selected samples to update the actor, the resulting policy displayed a more robust behaviour with respect to the imposed constraints.

The present work assumes a combination of DRL with a classical control method, defining a learning-based adaptive control system. In particular, we propose to work in the poles domain of the control law as it is easier there to define constraints for stability purpose compared to the space of gains. The high-dimensional continuous state and action spaces are successfully handled by the use of Maximum Entropy DRL \citep{Ahmed2019UnderstandingTI}, instead of using the DDPG or TD3 algorithms whose optimisation procedure is mostly based on the reward. The paradigm adopted in this paper helps the agent to build a more robust policy by forcing the exploration of suboptimal trajectories, resulting in improved generalisation capabilities. Process uncertainty is further taken into account by building a stochastic policy, in contrast to a deterministic one as done in other work. The effect of partial observability of the process is amplified since, in the present context, the current disturbance is not available. This issue is alleviated by considering an augmented state-space representation of the AUV process where the IMU feedback is incorporated into the state vector in such a way that it indirectly captures the effect of the disturbing forces, allowing the DRL algorithm to compensate for it. Finally, the Sim-to-Real Transfer of the policy is achieved by reducing the distribution-shift problem via an improved Domain Randomisation method \citep{Tobin2017DomainRF}. The strategy for incorporating DRL in the adaptive control framework developed in this work, aiming to design a learning-based adaptive controller, is presented as follows.

\section{Sim-to-Real Transfer of Adaptive Control
Parameters for AUV Stabilisation} \label{AUV}

The objective of this study is to propose an adaptive control architecture, combining DRL and model-based control, capable of adjusting its parameters to current disturbances that are not measured. We summarise in the following the key elements of this analysis.

\subsection{Task description} \label{task_description}
The problem domain considered in this work is the control of manoeuvring tasks for AUVs. The primary control objective is to achieve multi-station keeping, which entails stabilising the AUV successively at various spatial setpoints, each defined by a specific position and orientation, for a predetermined duration. The stabilisation is considered successful if, over a specific amount of time, the distance to the target position and orientation remains under a predefined threshold.

\subsection{Simulated and Real-Word Robotic Systems} \label{simulation-details}
For the concept of transfer learning to be demonstrated, both simulated and real-world testing platforms were used.  For this project, the chosen hardware platform was a modified Blue Robotics BlueROV 2 Heavy configuration \citep{bluerov2}.  The BlueROV is a low-cost compact Remotely Operated Vehicle (ROV) that has been applied in a variety of situations ranging from hobbyist use to applications in aquaculture and inspection of marine objects.  The heavy configuration adds two extra thrusters making the vehicle over-actuated, allowing control over all its six axes.\\
Flinders University's BlueROV has been modified to include an ALVAR AR tag \citep{alvar} for pose measurements, by an externally mounted camera, in addition to the acceleration and rotational rate from the Pixhawk Fight Controller and depth from the onboard pressure sensor. These measurements are fused by an EKF system estimating the BlueROV's 6 DoF pose and velocities, see Section \ref{subsec:positioning}.
This vehicle has also been modified to use the Robotics Operating System (ROS) \citep{Quigley2009} middle-ware. ROS is a very popular system in research robotics allowing the rapid development of robotic systems by the combination of small and simple programs called \textit{nodes} that communicate information via \textit{topics}.

\subsection{Evaluation}
Once trained in a simulated environment, the resulting policy was evaluated in the real world against its model-based counterpart defined in \cite{BluerovFLINDERS} which consists of a model-based PID controller with fixed gains determined using Robust Control theory \cite{CDoyle}. The evaluation was split into two scenarios: with and without varying current disturbances. Since both controllers were based on the exact same PID control structure, it is fair to compare them as they produce analogous control inputs. The detailed descriptions of the simulations and evaluation are given in Section \ref{Expe_description}.

The next section presents the control design starting with the model-based part of the proposed learning-based adaptive controller.

\section{Model-based control structure} \label{model-based}

\subsection{Model description} \label{description}

The state of the vehicle described in Section \ref{AUV} at the timestep $t$ denoted as $x_t$ is defined by its Cartesian position and Euler orientation $x_t=[x_t\ y_t\ z_t\ \psi_t\ \theta_t\ \phi_t]^T$ (respectively roll, pitch and yaw for its orientation). A setpoint is defined as $x_{w}=[x_w\ y_w\ z_w\ \phi_w]^T$ and the error on the setpoints is defined as $e=x_t-x_w$. Here, given the characteristics of the vehicle, the roll and pitch angle of the AUV are not controlled, as they need to vary in order for the vehicle to perform sway and surge movements, whereas these parameters stabilise to $0$ when performing station keeping given the centre of buoyancy of the AUV. The control objective is therefore to minimise the Euclidean distance $d_t$ between the AUV and the setpoint:
\begin{equation} \label{ctrl-obj-expe}
\begin{split}
d_t = \Big[(x_w-x_t)^2&+(y_w-y_t)^2\\
&+(z_w-z_t)^2+(\phi_w-\phi_t)^2\Big]^{\frac{1}{2}}.
\end{split}
\end{equation}
The task of station keeping can be achieved if the following control objective is met:
\begin{equation} \label{ctrl-objective-exp}
\forall\ t'\in[t-\varsigma,t],\ \nexists\ i \in\mathbb{R}^u \text{ such as }\ \mid e_i(t')\mid\ > d_{reached},
\end{equation}
where $d_{reached}$ is the predefined threshold value on the errors within which the setpoint is considered satisfied. This class of control objective is used in various AUV missions, such as autonomous docking or underwater inspection.

The physical model of the AUV, which encompasses the known part of the controller ($f_1$), can be summarised using the state-space representation described below \citep{Fossen1994NonlinearMA, YCMLW15}:
\begin{align} \label{state-maritime-vehicle}
&\dot{\eta}=J_\Theta(\eta)\nu, \ \\
&M\dot{\nu}+C(\nu)\nu+D(\nu)\nu+g(\eta)=\delta+\delta_{cable},
\end{align}
where $\eta$ and $\nu$ are position and velocity vectors, respectively, $\delta$ is the control force vector, and $\delta_{cable}$ represents the forces from the cable attached to the AUV. The BlueROV 2 Heavy is equipped with 6 thrusters and the control vector $u$ is obtained using the equation $\mathbf{\delta}=\mathbf{T}(\alpha)\mathbf{Ku}$, where $\mathbf{T}(\alpha)\in\mathbb{R}^{n\times r}$ is the thrust allocation matrix, $\mathbf{K}$ is the thrust coefficient matrix, $\mathbf{\delta}$ is the control force vector in $n$ degrees of freedom (DOF), and $\textbf{u}\in\mathbb{R}^r$ is the actuator input vector. The AUV is subject to an additive but unknown current disturbance which can be modelled as:
\begin{equation} \label{law:wind}
    u_{adp}= u + u_\mathrm{current}
\end{equation}
In this context, despite $u_{current}$ being unknown, the PID control law can be considered as the integral term ensures convergence to the steady state as $t\xrightarrow{}\infty$ despite the presence of the current disturbance.

\subsection{Adaptive Pole-placement strategy}

This work assumes that the state of the AUV is fully observable and controllable as we have sensors which provide measurements of its linear and angular velocities as well as its orientation in terms of Euler angles. However, only the AUV's inertial measurement unit (IMU) feedback is available, and the characteristics of the current disturbance cannot be directly measured or estimated. Consequently, a PID-type control law is considered as a baseline solution \citep{aastrom2021}.

The PID control law in state-space form is given by $\dot{X}=(A-BK)X$, and can be expressed as Eq.~\ref{cl-system}, where $k_p, k_i$, and $k_d\in\mathbb{R}^+$ are the control gains. Anti-windup is applied to the integral term $\sigma(t)=\int^t_0e(t)dt$, and a low-pass filter is used on the derivative term to reduce the oscillations caused by process noise. The resulting PID control law can be defined as follows:
\begin{equation}\label{cl-system}
\begin{split}
u(t)=\ &k_p e(t) + \min (k_i \sigma(t), u_{max}(t))\\
&+ k_d\left[ (1-r) E_{t-1} +  \frac{r de(t)}{dt}\right],
\end{split}
\end{equation}
where $u_{max}(t)$ is the maximum control input that can be sent to the AUV speed controllers, $|u(t)|<u_{max}(t)$, $r=e^{-T_s/T_f}=e^{-4}$ (\cite{DeLarminat}) is a smoothing factor, and $E_t$ is the output of the filter at the previous step. To ensure some stability of the control law in terms of output boundedness, the poles of Eq.~\eqref{cl-system} must be placed in the complex left half-plane. Candidates for eigenvalues are determined as solutions of $\lambda^3+\lambda^2k_d+\lambda k_p + k_i=0$. To maintain the gain space dimension, the pole-value candidates $\tau_i\in\mathbb{R}^+$ are defined as follows:
\begin{align} \label{poles}
\lambda_{1}= \cfrac{-1}{\tau_1}\ ;\ \lambda_{2}=\cfrac{-1}{\tau_2}\ ;\ \lambda_{3}=\cfrac{-1}{\tau_3}\ .
\end{align}

The controller gains are determined through a resolution and transformation process explained in detail in \citep{Chaffre2021DirectAP}. By considering the design in Eq.~\eqref{poles}, the bounds for the controller parameters can be defined based on control constraints that are easier to derive in the pole domain. In this case, for any $\tau_i>0$, the poles of the feedback loop are placed on the x-axis of the complex left half-plane \citep{Chaffre2021DirectAP}. Based on the control objective described in Eq.~\eqref{ctrl-objective-exp}, the desired maximum settling time of the closed-loop control is set to $\varsigma=10$ seconds, indicating the maximum time allowed for the system outputs to remain within $\chi=5\%$ of their desired values. According to the design we proposed in \cite{Chaffre2021DirectAP}, the final control input is given by:
\begin{equation}
\begin{split}
u(t)=\ &\cfrac{(\tau_1+\tau_2+\tau_3) e(t)}{\tau_1\tau_2\tau_3}\ \\
&+\min (\cfrac{\sigma(t)}{\tau_1\tau_2\tau_3}, u_{max}(t)) \\
&+\cfrac{\tau_1\tau_2 + \tau_1\tau_3 + \tau_2\tau_3}{\tau_1\tau_2\tau_3} \left[ (1-r) E_{t-1} + \frac{r de(t)}{dt}\right], \label{adpp-input}
\end{split}
\end{equation}
\textcolor{black}{where $\tau_1,\tau_2$ and $\tau_3$ are the poles of the considered feedback loop controller \citep{Chaffre2021DirectAP}.} In scenarios where only limited information about environmental disturbances is available and in the presence of time-varying processes, model-free adaptation can be employed. To account for the uncertainties in pole selection, the proposed approach uses DRL to construct a stochastic predictive model $\pi_\mu$. This model maps the AUV state vector $s_t$ to the pole values. The learning agent aims to build a predictive model that directly maps the AUV state to the pole values $\tau_i$, which are then used to compute the PID control inputs to regulate the AUV's position and orientations. This mapping \citep{Chaffre2021DirectAP} ensures that for any pole values chosen by the DNN in the parameter cube \citep{Chaffre2021DirectAP}, the resulting control input will maintain the closed-loop poles in the left half-plane. The stability of the control loop must also be taken into account when considering its transfer on actual AUVs, especially due to their substantial operating expenses and the elevated risk of vehicle loss in a real maritime environment. We have demonstrated in prior work \cite{HectorKohler} how Lyapunov stability analysis can be conducted for the proposed learning-based adaptive control design \eqref{adpp-input} in the context of AUVs.

\section{Model-free adjustment mechanism}

\subsection{Stochastic policy}

In order to take into account the uncertainties in the poles selection, we propose to use DRL to build a stochastic predictive model $\pi_\theta$ that maps a state vector $s_t$ into the pole values:
\begin{equation} \label{desired-policy-s4}
\left\{
\begin{aligned}
\pi_\theta: \hspace{.2cm}&S \subset \mathbb{R}^{dim(S)} &\rightarrow &\hspace{.2cm}A \subset \mathbb{R}^{3\times dim(u)}\\
&x=\left[\textbf{s}_t \right]^{T} & \mapsto& \hspace{.2cm} \left[\lambda_i,\mu_i\right].
\end{aligned}
\right.
\end{equation}
where $\mathcal{N}(\tau_i)$ is the probability density function of $\tau_i$ that is modelled by a Normal distribution $\mathcal{N}(\tau_i)$ as:
\begin{equation} \label{predictive-model}
\mathcal{N}(\tau_i)=(2\pi\mu_i)^{-1/2}\exp\{-\frac{1}{2\mu_i}(x-\lambda_i)^2\},
\end{equation}
with $\lambda_i\in\mathbb{R}$ and $\mu_i\in\mathbb{R}^+$ are the mean and variance of $p(\tau_i)$ that are estimated by the Policy network. Therefore, the outputs of the Policy network are the 3$\times dim(u)$ pairs of $(\lambda,\mu)$ representing the Normal distributions $\mathcal{N}(\tau_i)$ used to sample the poles for each control input $u_i$.
This stochastic policy Eq. \eqref{desired-policy-s4} prevents early convergence, encourages exploration, and improves the robustness to uncertainties.

In practice, the pole $\tau_i(t)$ is sampled from $\mathcal{N}(\tau_i)$ after applying an invertible squashing function (i.e. tanh) to $\mathcal{N}(\tau_i)$ (in order to bound the Gaussian distribution) and after using the change of variable to compute the likelihoods of the bounded action distribution (see Appendix C of \cite{haarnojaSoft1} for the complete description of this process). Designing this stochastic function \eqref{desired-policy-s4} is numerically expensive due to the dimensions of the underlying spaces, excluding real-time computation with model-based methods only. The DRL framework allows us to iteratively build an estimate of this optimal mapping function.

\subsection{State vector} \label{state-vec}

In this work, at each timestep, the agent captures an observation vector $o_t$ representing the process dynamics which consists solely of variables that are available on the real vehicle. The observation vector is thus defined as:
\begin{equation}\label{obs-vec-expe}
o_t=[\ a_{t-1}\ ;\ \Theta\ ;\ V\ ;\ \dot{V}\ ;\ \Omega\ ;\ e_t\ ;\ e_{L2}\ ],
\end{equation}

where
\begin{itemize}
    \item $a_{t-1}\in\mathbb{R}^{18}$ are the last actions estimated (i.e. poles value),
    \item $\Theta=[\phi;\theta;\psi]$ are the Euler orientation of the vehicle (roll, pitch, and yaw respectively),
    \item $V=[v_x;v_y;v_z]$ and $\Omega=[\omega_\phi;\omega_\theta;\omega_\psi]$ are respectively the vehicle's linear and angular velocities,
    \item $\dot{V}=[\dot{v_x};\dot{v_y};\dot{v_z}]$ is the vehicle's linear acceleration,
    \item $e_t\in\mathbb{R}^6$ are the error values on each setpoint as defined in Section \ref{description},
    \item and $e_{L2}$ is the Euclidean distance to the steady-state defined as $e_{L_2} = \sqrt{\sum_{i=1}^{i=dim(u)} e_i^2(t)}$.
\end{itemize}

The dimension of the observation vector $o_t$ is therefore equal to 40. It is important to note that with this observation vector (Eq.~\eqref{obs-vec-expe}) the current disturbance characteristics are not included. To improve the observability of the process and following our previous results~\citep{chaffre:hal-02958155}, the state vector $s_t$ is obtained out of the current and past observation vectors along with their two-by-two difference. This results in a 120-dimensional state space defined as:
\begin{equation}
    s_t=[o_t\ ;\ o_{t-1}\ ;\ o_{t-1}-o_t] \in \mathbb{R}^{120}.
\end{equation}

DDPG algorithms have shown promise in handling the control tasks of real-world systems \citep{ye2021survey}. This architecture simultaneously estimates a value function and a policy function to improve the agent's performance. Off-policy methods, using Experience Replay (ER), have been developed to enhance the sample efficiency of these functions using past experiences generated by different policies. However, a critical challenge faced by DDPG and TD3 algorithms is the value overestimation problem \citep{Kumar2019StabilizingOQ}. To mitigate this, we use the Maximum Entropy DRL algorithm named SAC \citep{Haarnoja2018SoftAA} which provides more robust and efficient learning in DRL-based control systems. Next, we introduce the version of the SAC algorithm used in this work.

\subsection{Soft Actor-Critic with Automatically Adjusted Temperature}

The Soft Actor-Critic (SAC) algorithm \citep{Haarnoja2018SoftAA} is a Deep Policy Gradient method known for its robustness to uncertainty and its suitability for operating in partially observable processes. SAC combines three key components: improved exploration and stability through entropy maximisation, an Actor-Critic architecture with separate Q-Value and Policy networks, and an off-policy formulation using Experience Replay. The objective function of the original SAC algorithm is defined as:
\begin{equation} \label{SAC-ori}
J(\pi|s_0)=\max_\pi\mathbb{E}_\pi[\sum_t r(s_t,a_t) + \alpha \mathcal{H}(\pi(\cdot|s_t))\big|s_0],
\end{equation}
Unlike traditional reinforcement learning algorithms that optimise the expected sum of rewards, SAC  also maximises the entropy $\mathcal{H}(x)=\mathbb{E}_{x\sim P}[-\log{(x)}]$ of the behaviour policy weighted by a temperature parameter $\alpha$ (cf. Eq. \eqref{SAC-ori}). This encourages exploration and flexibility in the agent's actions, making it more robust to variations in the environment, by forcing the agent to explore sub-optimal trajectories until the optimal ones for the long-term objectives are eventually observed. The entropy term is explicitly incorporated into the State-Value function, which combines Q-Value estimates and the policy's entropy. The temperature parameter $\alpha$, which controls the stochasticity of the policy, is here controlled indirectly by the reward scale. Unfortunately, it is often difficult to define beforehand an optimal reward scale as the entropy can vary unpredictably both across tasks and during training as the policy becomes better. To cope with this limitation, in this study we used an improved version of the SAC algorithm \citep{haarnojaSoft3} which consists in adapting the temperature term so as to maintain the desired entropy value:
\begin{equation} \label{dual-obj}
    \max_{\pi_0, \dots, \pi_T} \mathbb{E} [ \sum_{t=0}^T r(s_t, a_t)] \text{ s.t. } \forall t\text{, } \mathcal{H}(\pi_t) \geq \mathcal{H}_0.
\end{equation}
To reduce value overestimation, this version of SAC \citep{haarnojaSoft3} utilizes two Q-Value function estimators and applies TD-Learning to iteratively estimate the Q-Value functions. The State-Value function $V(s_t)$ is not explicitly represented anymore by an DNN, but it is implicitly defined through the Q-Value functions and the policy (as no differences are observed when comparing both methodologies \citep{haarnojaSoft3}). The delayed update trick from the TD3 algorithm is also employed \citep{Fujimoto2018AddressingFA}, which limits the likelihood of repeating updates with respect to an unchanged critic, limiting the variance of the value estimate, resulting in higher quality policy updates.

The policy network's parameters are consequently optimised to minimise the expected Kullback-Leibler divergence between the current policy and the exponential of the Q-Value function. With this version of SAC, the reward scale does not need to be tuned as the relative weight of the entropy term is adapted to satisfy a minimal entropy constraint. The resulting dual constraint optimisation for the policy can be defined as:
\begin{equation} \label{obj-alpha-bluerov}
J(\alpha)=\mathbb{E}_{s_t\sim D,a_t\sim\pi_\mu}\big[-\alpha\log{\pi_\mu(a_t|s_t)}+\alpha \times 18 \big].
\end{equation}
with $\mathcal{H}=-18=-dim(u)$ is the target entropy, which according to \cite{haarnojaSoft3} can be easily set to the dimension of the action space. In the present case, the action space has a dimension of 18 since the task is to control a vehicle with 6 degrees of freedom (DoFs), and each DoF has 3 possible pole values.

\subsection{Reward function}

Since we are using the second version of SAC \citep{haarnojaSoft3}, the reward scale does not require to be tuned. Thus, we proposed the following reward design:
\begin{equation} \label{reward-expe}
r(s_t)=\exp\left[-(e_{L2}(t)) \right].
\end{equation}
This reward function is solely a function of the distance to the setpoint included in the state vector in Eq.~\eqref{state-vec} which can be maximised if and only if the desired distance (Eq.~\eqref{ctrl-obj-expe}) is minimised. With this second version of the SAC algorithm \citep{haarnojaSoft3}, the reward scale does not need to be controlled. Therefore our reward signal \eqref{reward-expe} is defined as $r(s_t)\in[0,1]$ which is more appropriate when using DNNs. The Soft Actor-Critic method, summarised above, was extended with the Biologically-Inspired Experience Replay (BIER) method introduced in the next section.

\subsection{Biologically-Inspired Experience Replay (BIER)}

The BIER method \citep{CHAFFRE2022} aims to combine the resilience of on-policy sampling with the data efficiency of off-policy formulation and, in general terms, it is defined by two distinct memory units: the sequential-partial memory (B1) and the optimistic memory (B2).

The B1 memory unit serves a similar purpose as the memory buffer in the original definition of Experience Replay (ER). In the context of robotics, where optimal behaviour is often highly temporally correlated, learning a limited set of such sequences can efficiently lead to optimal behaviour. 
However, using temporally correlated samples can compromise learning in DNNs of the underlying SAC method due to overfitting and lack of independence and identical distribution (I.I.D.) in the training dataset. To address this issue, BIER incorporates the concept of partial transitions in B1, whereby only one out of every two transitions is added to this buffer. This approach adds a regularisation effect to the DNN fitting process, reducing the age of the oldest policy stored in B1, thereby improving the learning performance \citep{Fedus2020RevisitingFO}.

The B2 memory unit represents an optimistic memory and is inspired by the observation that positive reinforcement is more efficient in biological systems than a combination of positive and negative rewards. B2 stores the upper outliers of the reward distribution, which are considered to be the best transitions. By increasing the probability of using past transitions associated with high-quality regions in the solution space, B2 aims to enhance performance improvement \citep{Fedus2020RevisitingFO}. 

Finally, BIER consists of randomly sampling $n$ temporally correlated sequences from B1 (i.e. a temporal sequence composed of $n$ consecutive transitions) and randomly sampling $n$ uncorrelated transitions from B2 to construct the mini-batch of past experience to perform the mini-batch gradient descent optimisation procedure of the DNNs.

\subsection{Domain randomisation} \label{DR-expe}

Despite the stability components of our learning-based adaptive controller described in \citep{Chaffre2021DirectAP}, training directly on the real platform is not a possibility due to the vehicle's limited battery life, added to the time to run the number of trials needed to train the ML models. Therefore, in this work, training was performed on a simulated version of the BlueROV platform, and the learned policy was transferred to the physical platform.  In this case, the distribution shift arises from the transfer of a policy trained in a near-perfect state space (obtained in a simulated environment) to an agent subject to sensor noise, delays, and a real turbulent environment.

Various techniques exist to reduce the reality gap between simulation and the real world, such as Domain Randomisation (DR) \citep{Tobin2017DomainRF}. In DR, the environment used for training is referred to as the \textit{source domain}, while the environment we aim to transfer to is denoted as the \textit{target domain}. Typically, training is only feasible within the source domain, where a set of $N$ randomisation parameters can be modulated to alter the domain's characteristics. Thus, a configuration $\xi$ can be defined as a sample drawn from a randomisation space $\xi\in\Xi\subset\mathbb{R}^N$. During training, data from the source domain are collected with the application of randomisation to the parameters. By doing so, the policy is exposed to a diverse range of distinct versions of the source environment, allowing for a better generalisation to be learned compared to exposure to a single environment. The appearance of the environment can be controlled by the following randomisation parameters: position, shape, and colours of the objects; the texture of material; lighting condition; random noise added to images; or position, orientation, and field of view (FoV) of the simulated camera. These parameters can also control the physical dynamics of the domain such as: mass and dimension of objects; mass and dimension of vehicles; damping and friction of the joints; observation noise, or action delay.

The idea of incremental environment complexity \citep{chaffre:hal-02958155} was employed in this study as a modification of the DR procedure. The approach involved training the agent in diverse variations of the same environment, each differing in task complexity as indicated by the quantity and shape of obstacles present. The agent would transition between these domains based on its performance, as assessed by the success rate. This method offers the advantage of preventing the agent from becoming trapped in an unfavourable regime by returning it to a previously solved complexity level if it fails to solve the current one. By appropriately adjusting the parameters, a smooth transition can be ensured as the agent progresses through each configuration until reaching the final one. However, it is important to note that this approach lacks control over the amount of data collected from each complexity configuration. Consequently, some configurations may be extensively explored while others receive less attention, potentially leading to overfitting. We can mitigate this issue by forcing the agent to collect the same amount of data from each complexity configuration, from the simplest to the more challenging one \citep{Chaffre2021LearningbasedVM}. Nevertheless, it is difficult to determine beforehand the appropriate amount of data that the agent will require to solve a configuration. With this approach \citep{Chaffre2021LearningbasedVM}, additional tuning of this parameter is necessary to ensure that no time is wasted on already solved configurations and that enough time is provided on the difficult ones.

For these reasons, the following three complexity configurations were considered in this work, as measured by the amount of disturbance:
\begin{itemize}
    \item Configuration 1: no disturbance at all.
    \item Configuration 2: current disturbance that does not vary within the episode.
    \item Configuration 3: current disturbance that changes at a random time within the episode between timestep 100 and 400, out of 500. The value 500 was chosen as the maximum value for the length of the episodes in accordance with the desired settling time defined in Section \ref{task-descrip}.
\end{itemize}

\begin{figure}[htbp]
\centering
\includegraphics[width=0.9\linewidth]{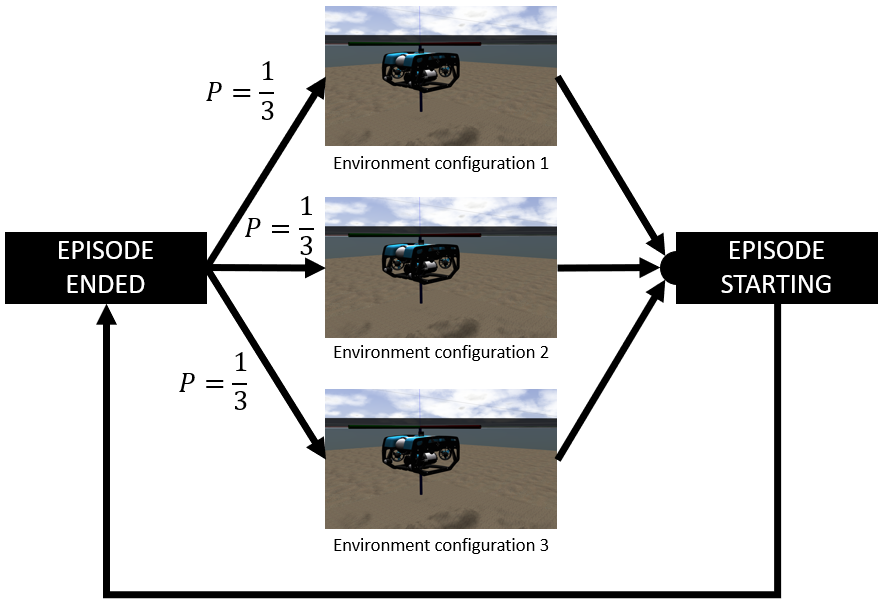}
\captionsetup{justification=justified}
\caption{Illustration of the domain randomisation technique. During training, the agent experienced a large number of variations of 3 environment configurations. Each configuration has the same probability $P=1/3$ to be chosen.}
\label{domain-rando-exp}
\end{figure}

Finally, before the beginning of each episode, the agent has an equal probability of experiencing each complexity configuration $P$. By doing so, we are sure that each configuration will be explored uniformly (to avoid overfitting) and the hardest configuration of the environment, which is the closest to the target domain, will be experienced very early in the training phase (for improved sample efficiency). This methodology is illustrated in Figure \ref{domain-rando-exp} where the choice of complexity configuration is performed after the end of each episode.

\subsection*{Exploration strategy} \label{Explo}

We used adaptive parameter noise \citep{Plappert2018ParameterSN} where random Gaussian noise $\mathcal{N}(0,\sigma)$ is added to the parameters of the policy network (weights and bias) at the beginning of each episode (then kept during the rollout) as:
\begin{equation}\label{param-noise1}
\sigma_{k+1}=
\left\{
\begin{array}{l}
\alpha\sigma_k,\hspace{2mm}\textbf{ if }d(\pi,\Tilde{\pi})<\delta,\\
\frac{1}{\alpha}\sigma_k,\textbf{ otherwise}.
\end{array}
\right.
\end{equation}
The noise standard deviation $\sigma$ is adapted according to a distance measure $d(\cdot)$ between the non-perturbed $\pi$ and perturbed policy $\Tilde{\pi}$ defined in~\cite{Plappert2018ParameterSN} as:
\begin{equation}
d(\pi,\Tilde{\pi})=\sqrt{\frac{1}{N}\sum_{i=1}^N\mathbb{E}_s[(\Tilde{\pi}(s)_i-\pi(s)_i)^2]},
\end{equation}
where the metric $\mathbb{E}_s[\cdot]$ is estimated over a batch of $1000$ samples from the Replay Buffer, the initial standard deviation is set to $1.0$, the threshold is set to $0.10$ and the update rate is set to $\alpha=1.005$. This strategy can be seen as a middle ground between evolution strategies and DRL. As recommended by the authors \cite{Plappert2018ParameterSN}, in order to avoid local maxima, which can still happen with a perturbed and stochastic policy, the parameter noise was combined with an $\epsilon$-greedy policy where each action holds an independent probability $\epsilon=0.01$ to be random.

\section{Learning-based Adaptive Pole-Placement}

\begin{figure}[htbp]
\includegraphics[width=\linewidth]{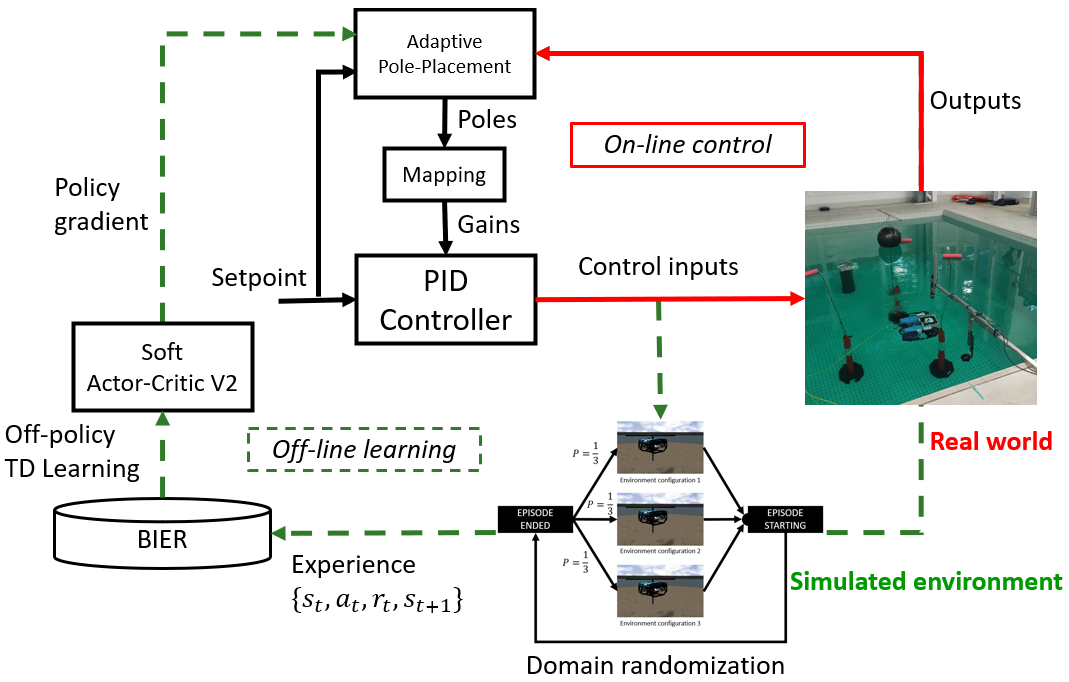}
\caption{Illustration of the overall proposed learning-based adaptive control system.}
\label{lb-ac}
\end{figure}

Figure \ref{lb-ac} summarises the overall learning-based adaptive control methodology proposed in this paper. We designed an adaptive pole-placement control structure where the gains of the PID law are transformed into the poles domain to be placed in appropriate locations by a DRL-based policy, before being transformed back into the temporal domain to compute the associated PID control input. These pole values are estimated by a policy learned by a DRL-based policy. In our case, we used the second version of the SAC algorithm \citep{{haarnojaSoft3}} to learn the optimal policy with respect to the considered reward function Eq.~\eqref{reward-expe}. The policy, along with the value functions, is learned offline with TD Learning \citep{{Sutton-Barto2ed}} using the BIER method \citep{CHAFFRE2022} for improved sample efficiency and by using the improved domain randomisation methodology defined in Section \ref{DR-expe}. After training, the resulting policy is directly transferred to the real platform. 

\section{Simulated training}

The simulation environment was based on the Gazebo robotics simulator and used the UUV Simulator package \citep{Manhaes_2016}. This combination provided a number of modules implementing a variety of maritime systems including a range of maritime sensors and systems.  The vehicle hydrodynamic model was based on the work of \cite{wu20186} who developed a model of the vehicle based on a combination of theoretical work and data published by \cite{sandoy2016system}.   

The simulation was then configured using the mass, added mass, linear and quadratic drag terms derived from \cite{wu20186}. The ROVs thrusters were implemented using the simple first-order thruster model from UUV Simulator, and placed on the model with the orientations and moments as estimated by \cite{wu20186}. The desired force and torque of the simulated vehicle were set by a ROS topic, and individual thruster effort was allocated using the Thruster Allocation Matrix (TAM) provided by the simulator.  The vehicle's pose and velocity estimates were published as a single Odometry message on another ROS topic.  Using these topics, a MIMO control system was used to guide the vehicle to perform station keeping.

A simulated training episode was defined as follows:
\begin{enumerate}[label=(\arabic*)]
    \item At the beginning of the episode the AUV was initialised at the position $(x_0,y_0)\in[-5,5]$, $z_0\in[-20,-10]$ with null velocity and a random orientation $(\psi_0,\theta_0,\phi_0)\in[\frac{-\pi}{4};\frac{\pi}{4}]$.
    \item A random configuration of the environment was generated as defined in Section \ref{DR-expe}.
    \item A random setpoint was generated with coordinates defined as: $(x_w,y_w)\in[-5,5]$, $z_w\in[-15,-5]$, $(\psi_w,\theta_w)=0$, and $\phi_w\in[-\pi/2,\pi/2]$.
    \item Then, the off-policy exploration strategy was used and the episode ended when the step number reached $500$.
\end{enumerate}
The training consisted of performing a total of 5000 episodes of maximum timesteps set to 500 which took approximately 4 hours (considering that the training was conducted using a "real-time factor", this implies that it is equivalent to four hours of actual vehicle usage in real-life conditions). Before an episode begins, the configuration of the environment characteristics was chosen as described in Section \ref{DR-expe}.

The complete list of hyperparameters is provided in Table \ref{hyperparam-expe} with the details on the DRL framework.

\begin{table}[htbp]
\scriptsize
\centering
\caption{List of hyperparameters and their values for experimental validation.\label{hyperparam-expe}}\vspace{0.25cm}
{\renewcommand{\arraystretch}{1}
\begin{tabular}{|p{4.25cm}|p{3cm}|}
\hline
\textbf{Training hyperparameter} & \textbf{Value} \\
\hline \hline
Number of hidden layers (all networks) & $2$ \\
Number of hidden nodes (all networks) & $256$ \\
Activation function & Leaky ReLU \\
Optimiser (all networks) & Adam \citep{Kingma2015AdamAM} \\
Learning rate (all networks) & $3\times10^{-4}$ \\
Discount factor ($\gamma$) & $0.99$ \\
Mini-batch size & $256$ \\
Target network smoothing coefficient ($\Delta$) & $0.005$ \\
Delayed update trick \citep{Fujimoto2018AddressingFA} & True \\
Critics L2 regularisation & $0.001$ \\
Layer Normalisation \citep{Ba2016LayerN} & True \\
Automatic temperature adjustment & True \\
Target entropy & $-18$ \\
Replay buffer max size & $1e6$ \\
Replay start size & $1e4$ \\
Experience Replay method & BIER\\
\hline
\end{tabular}}
\end{table}

\begin{figure}[htbp]
\centering
\includegraphics[width=\linewidth]{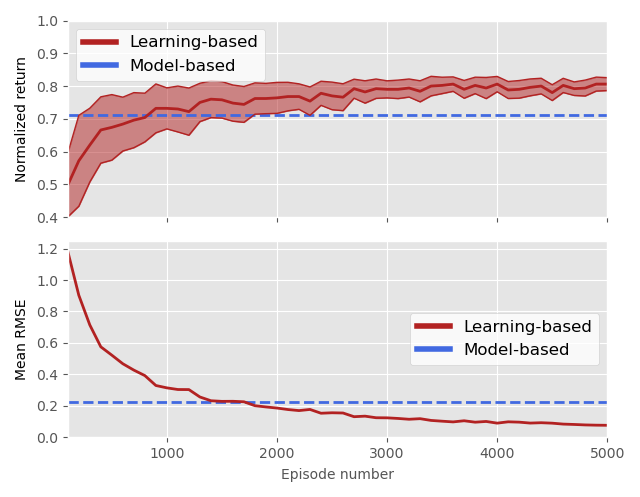}
\caption{Training curves of SAC with learned temperature. This process corresponds to simulated pre-training on 2.5 million samples, taking around 4 hours.}
\label{training-curves}
\end{figure}

The training curves are presented in Figure \ref{training-curves}. The performance of the proposed learning-based controller is depicted in red (with the shaded regions representing the standard deviation) and the performance of its model-based counterpart is represented in blue. These performances are the mean values of the aforementioned metrics computed over 100 episodes over a moving window of 100 episodes. As we can see in the top plot of Figure \ref{training-curves}, the learning-based controller was able to learn the task and converge toward what seems to be a maximum value of the reward. In the second plot of Figure \ref{training-curves}, the control performance is displayed in terms of RMSE on the setpoint. We can see that the learning-based adaptive controller exceeded the control performance of the model-based controller, which is represented by the blue horizontal lines. In the next section, we present the experimental evaluation of the resulting policy.

\section{Experimental setup} \label{Expe_description}

This section presents the results of the experimental evaluation campaign where the policy trained under simulation is 
transferred on a real vehicle 
in the environment depicted in Figure~\ref{experiments-settings-1}. This campaign covered approximately 280 minutes (or $\sim$4h40) of real-life operating time.

\subsection{Physical Vehicle}
\begin{figure}[htbp]
\centering
\captionsetup{justification=justified}
\includegraphics[width=\linewidth]{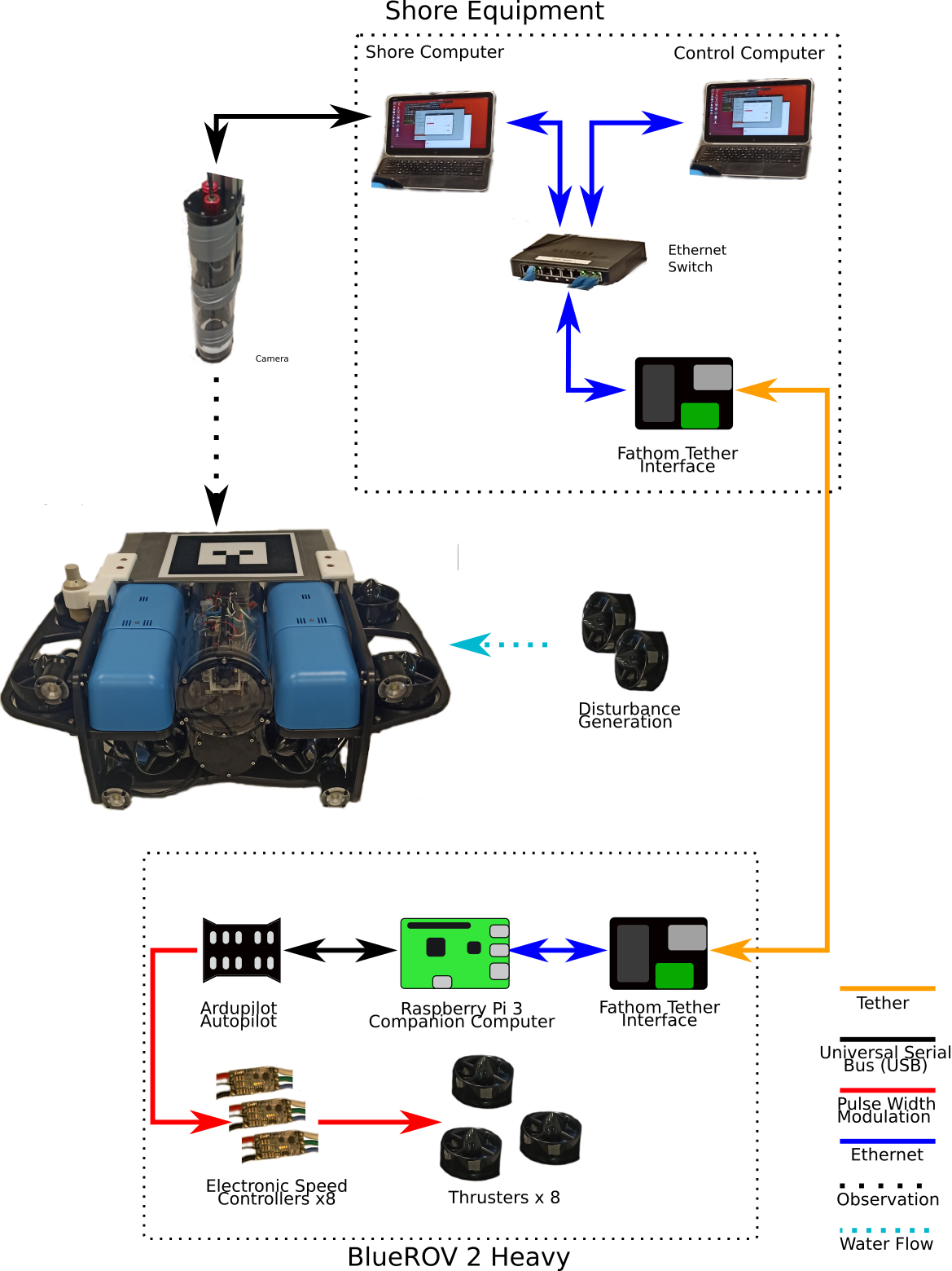} 
\caption{Physical block diagram of the experimental setup.  This diagram includes both the BlueROV 2 Heavy and the shore equipment used for monitoring and control.}
\label{bluerov-system}
\end{figure}

For the effective sim-to-real transfer, the physical 
vehicle should match the interface of the simulation. To meet this requirement, the system shown in Figure \ref{bluerov-system} was defined. It uses an Ethernet-based network, allowing the transfer of pose and control information between shore systems, while a pair of Blue Robotics Fathom-X boards allows Ethernet communication with the AUV across a tether. With this network, high-speed low-latency communication can be performed between the shore systems and the AUVs on-board computer systems. Combined with the ROS' ability to operate in a network transparent manner, robotics software can be distributed across multiple systems while performing effective estimation and control of the AUV.

The onboard processing on the AUV was provided by a Raspberry Pi 3 single-board companion computer.  This system ran an Ubuntu-based system with a ROS Kinetic package developed by Blue Robotics \citep{bluerovros}.  This computer communicated with a Pixhawk autopilot \citep{pixhawk} running the Ardupilot firmware via the Micro Aero Vehicle Link (MAVLink) protocol.  This allowed the control and monitoring of the vehicle using QGroundcontrol, a standard base station for drone vehicles \citep{qgroundcontrol}.  This data was also communicated to the ROS system using an instance of the MAVROS MAVLink to ROS gateway \citep{mavros}.

The BlueROV's PixHawk autopilot contains sensors capable of estimating attitude and orientation information but it is not capable of producing an absolute position estimate. Therefore, this work used a hybrid localisation system, with an external camera and a marker placed on the top side of the vehicle.

The camera was mounted within a Blue Robotics  waterproof enclosure with its optical centre in a transparent dome.  The camera assembly was attached to an aluminium frame such that the end of the enclosure was beneath the water level, and facing downward as illustrated in Figure \ref{cam-1}.  This gave a clear view of the BlueROV while minimising distortion due to refraction.  With the assembly mounted in the immersion tank, the camera was calibrated using a waterproof checker board.

In this configuration, the camera system had a clear view of the vehicle allowing visual tracking to be performed.  The marker pose was estimated using the ar\_track\_alvar ROS package \citep{artrackalvar} which uses the ALVAR library \citep{alvar} to track fiducial markers. The recovered pose of the marker was fused with data from the autopilot using the robot\_localization package \citep{MooreStouchKeneralizedEkf2014}.  This solution allowed a bounded estimate of vehicle pose and velocity information which was published as a ROS Odometry message of the same type as published by the simulator.

The control of the vehicle was done via a ROS Wrench message, containing both force and torque terms.  This information was mapped via a custom node into a joystick override message.  Estimation of the thruster effect is a challenging task, with the force generated by a thruster varying significantly based on factors such as the speed of the propeller, and the rate of advance of the vehicle.  For this project, the thruster effect was estimated from the force curves published by Blue Robotics \citep{t200}.  These are static thrust curves generated by bollard pull tests within a static environment.  Using these thrust curves, the thruster forces were linearised around the expected operating point of 1400-1600 microseconds, corresponding to a force of +/- 5 Newtons.  This was then used to generate the coefficients shown in Table \ref{joystick-mappings}.  This mapping was used to convert between the desired force on the vehicle and the control values passed to the vehicle autopilot.\\
\begin{table}[htbp]
\scriptsize
\centering
\caption{Wrench to joystick override mappings.}
 \begin{tabular}{c c c} 
 mapping & calculation & coefficient\\
 \hline
 thruster force & $25/(0.2 * 9.8)$ & 12.8\\
 roll torque & $thruster force/(0.218 * 4)$ & 14.7\\
 pitch torque & $thruster force/(0.12* 4)$ & 26.7\\
 roll torque & $thruster force/(0.1888 * 4)$ & 17.0\\
\end{tabular}
\vspace*{0.25cm}
\label{joystick-mappings}
\end{table}

Once in a suitable format, the message was sent to the autopilot via the MAVROS node.  The autopilot used the override signal with a TAM matrix to allocate effort to the vehicle's brushless Electronic Speed Controllers (ESCs), these controllers in turn drove the T200 thrusters that move the vehicle.

Using this configuration, the control system was tested on a real-world underwater vehicle using the same topics and message types as the simulated vehicle. In the next Section, further details on the positioning system are provided.

\subsection{Positioning system}\label{subsec:positioning}
As previously introduced, a positioning system was utilised to provide a continuous and accurate estimate of the AUV's pose and velocities in 6DOF as detailed in Eq.~\ref{state-maritime-vehicle}.
This estimate was generated by fusing the available measurements utilising an Extended Kalman Filter (EKF). Measurements in the configuration utilised in this experiment included acceleration and rotational rate from the IMU on the Pixhawk, depth from the BlueROVs pressure sensor, and pose from a tag tracking system utilising a webcam.
The tracking system utilised was the ar\_track\_alvar ROS package \citep{artrackalvar} which uses the ALVAR library \citep{alvar} to track fiducial markers in this experiment using a Microsoft Lifecam configured to the resolution of 720x1280 at a rate of 30 Hz in a waterproof housing. To account for the optical qualities of water the web camera intrinsics were calibrated in the configuration it was to be used in the experiment, i.e. underwater, at ranges anticipated for the experiment and consistent with ranges where tracking is possible.

\begin{figure}[htbp]
\centering
\includegraphics[width=0.9\linewidth]{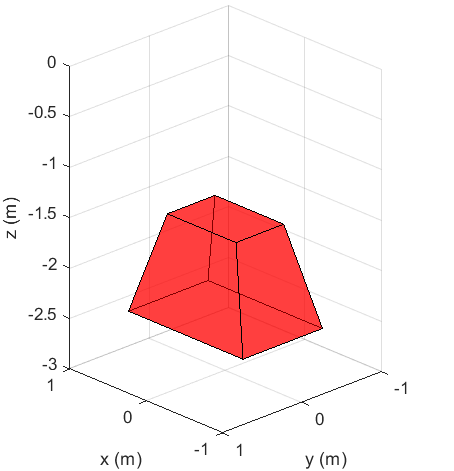}
\caption{Operational region of AR tracking system due to limitations in the camera lens calibration FoV and tracking.} \vspace{-0.125cm}
\label{posi-track-bounds}
\end{figure}

The fiducial marker was made as large as possible to maximise the range at which tracking could occur but was within the limits of the ROV. The tag was manufactured from laser-cut acrylic and treated to make the surface matte to prevent reflections. This resulted in a system which could track the marker at a distance between 0.5m to 2.5m of the camera. Due to the limitations in the camera's lens calibration, the tracking was optimally calibrated for a distance of 1.5 to 2.5m. The limitations in calibration in combination to the FOV of the camera and lens properties an optimal operating region was calculated, as illustrated in Figure \ref{posi-track-bounds}.

The measurements generated by the tracking system were transformed from the frame of reference of the camera to the frame of reference of the BlueROV making it suitable for integration in the state estimation.
\par
The above measurements were fused using an EKF implemented in the robot\_localization package \citep{MooreStouchKeneralizedEkf2014}. To maximise the responsiveness of the estimate the update rate of the implemented EKF was set to 50 Hz to match the data rate of the IMU, the fastest sensor.
The specific configuration of the measurements is presented in Table \ref{measurement-mapping} detailing the mapping between the sensors and the estimated state. The measurements of $z$, $\phi$, $\theta$ provided by the tag tracker were set differential, i.e. difference in the available measurements to generate rate measurements, to avoid inconsistencies, and biases, between these measurements and the more accurate measurements from the accelerometer and depth sensors for these states.

\begin{table}[h]
\scriptsize
\centering
\caption{Mapping of measurements to EKF.}
\begin{tabular}{c c c c} 
sensor        & rate (Hz) & state      & mapping\\
\hline
accelerometer & 50 & $\phi$            & absolute\\
              &    & $\theta$          & absolute\\
gyroscope     &    & $\omega_{\psi}$   & absolute\\
              &    & $\omega_{\theta}$ & absolute\\
              &    & $\omega_{\phi}$   & absolute\\
pressure      & 10 & $z$               & absolute\\
tag           & 30 & $x$               & absolute\\
              &    & $y$               & absolute\\
              &    & $z$               & differential\\
              &    & $\phi$            & differential\\
              &    & $\theta$          & differential\\
              &    & $\psi$            & absolute\\

\end{tabular}
\vspace*{0.25cm}
\label{measurement-mapping}
\end{table}

Using the above configuration, beyond the otherwise default configuration, the robot\_localisation package produced stable estimates, with approximate confidence bounds of +/- 5mm. Notable divergences from this behaviour occurred in scenarios corresponding to limits in the camera lens calibration and tracking system. When the depth of the AUV exceeded 2.5m, or when the tag went out of field view, tracking became inconsistent affecting the accuracy of the estimate.

\begin{figure}[htbp]
\centering
\includegraphics[width=0.9\linewidth]{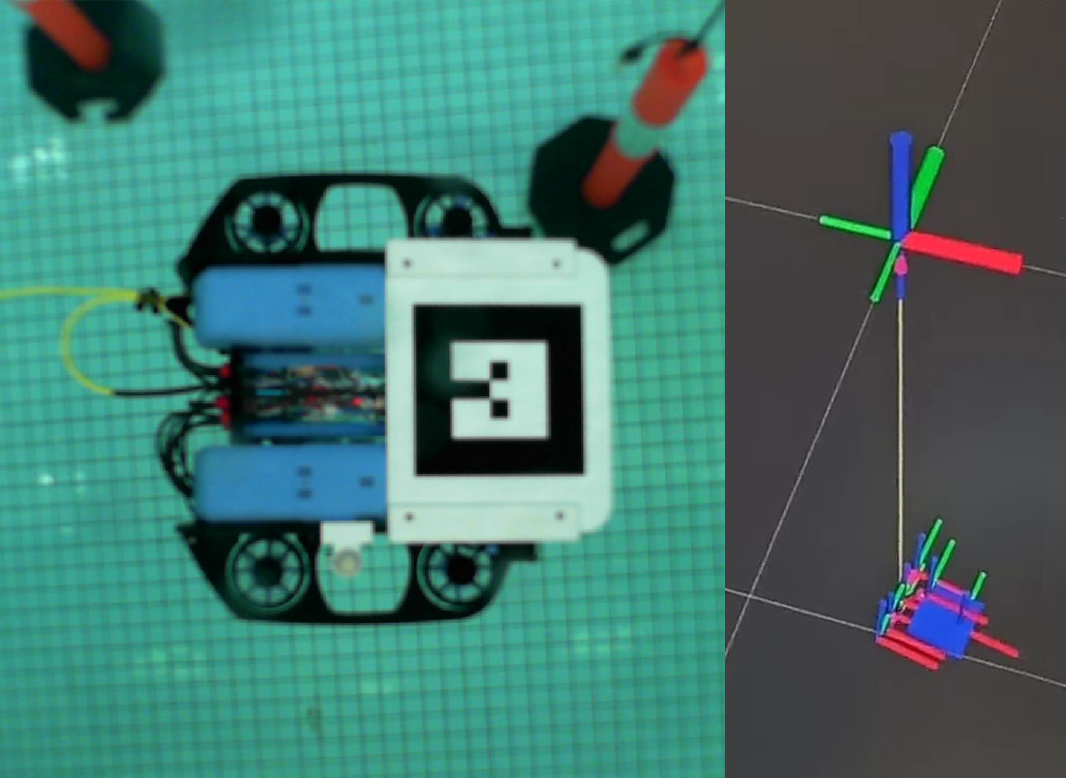}
\caption{Illustration of the camera feedback (left of the screen) and the EKF pose estimate represented by the position vectors (right of the screen). 
}
\label{posi-track-system}
\end{figure}

\subsection{Disturbance generator}
\begin{figure}[htbp]
    \centering
    \subfloat[\centering ]{{\includegraphics[width=5.140cm]{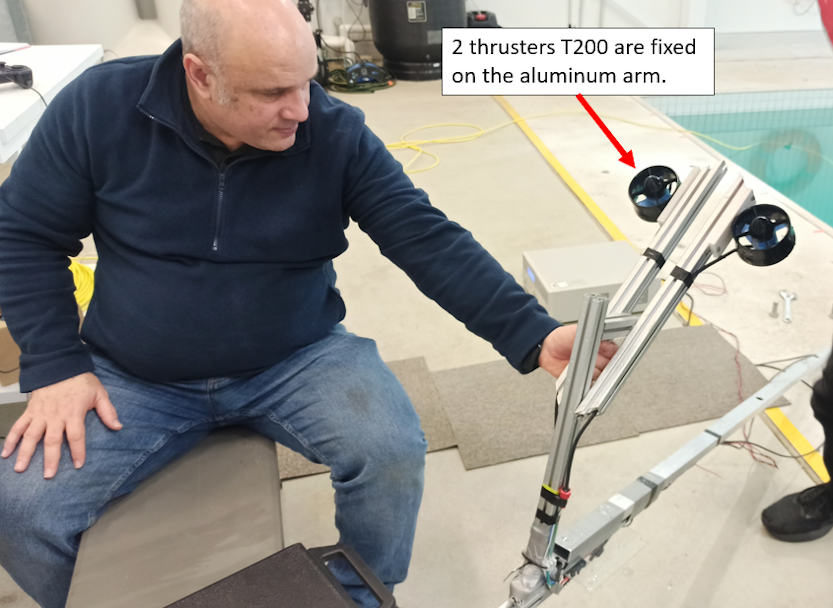} }}%
    \subfloat[\centering ]{{\includegraphics[width=2.110cm]{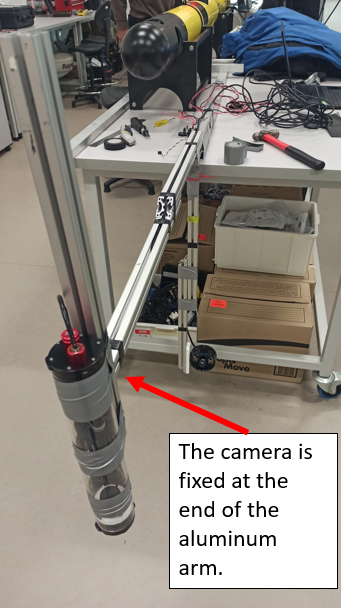} }}%
    \caption{Illustration of the disturbance generator system (a) and the marker tracking system (b).}%
    \label{dist-generator}%
\end{figure}

In order to evaluate the robustness of the controllers against disturbance, we proposed to create an artificial current in the water tank. To that end, we fixed two thrusters of type T200 (the same as the ones on the Bluerov platform) on the aluminium arm where the camera is attached as illustrated in Figure \ref{dist-generator}. We chose a particular placement and orientation of the thruster such as to optimise the field of effect in the pool. The thrusters are controlled through ESC input that we set to 1625, which according to Blue Robotics documentation gives around 8 Newtons of thrust per thruster. The total current draw for the pair is approximately 2.7A, providing a power draw of around 38 Watts.

\subsection{Task execution}\label{task-descrip}

\begin{figure}[htbp]
\centering
\includegraphics[width=0.9\linewidth]{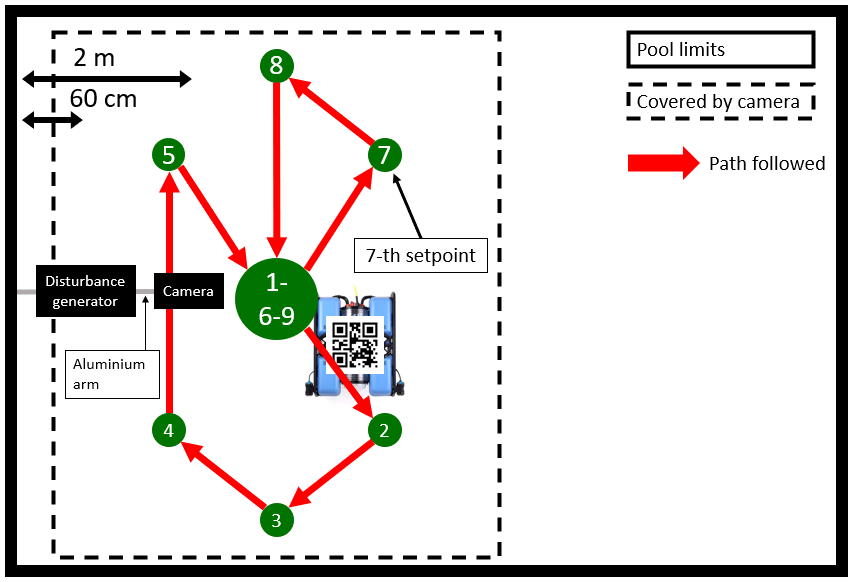}
\caption{Illustration of the multi-station keeping task performed during the experimental evaluation.
}
\label{exp-task}
\end{figure}

For the physical robot, the multi-station keeping control was executed as follows: starting from an initial position, the vehicle was required to perform station keeping for an amount of 1000 timesteps ($\sim$45 seconds) at each setpoint, as shown in Figure~\ref{exp-task}. Each session was, therefore, equivalent to $\sim7$ minutes. Both controllers were evaluated with respect to two environmental conditions: with and without current disturbance. Each station-keeping experiment was conducted 10 times by the physical robot for every control method examined in this study, as well as for each disturbance configuration. The reported results correspond to the average values obtained from these 10 trials.

\newpage
To mitigate potential bias in this evaluation, an initialisation procedure was executed wherein the vehicle is initially brought to the first setpoint. This involves activating the control system when the vehicle is in close proximity to the setpoint, allowing the control system to stabilise the vehicle over a fixed number of timesteps matching the duration of the subsequent experiments (i.e., 1000 timesteps). Subsequently, the experimental session commences from this same setpoint. In practice, it means that there is an additional setpoint 1 in the list presented in Table \ref{setpoints-list}, which is not taken into account when calculating the performance metrics. This guarantees that the AUV always starts around the same position. We found this practice to be particularly relevant when current disturbances were present since, without the additional setpoint, the starting point of the sessions was distinct due to the current-generated drift.
\begin{table}[htbp]
\scriptsize
\centering
 \begin{tabular}{|cccccccccc|} 
 \hline
Setpoint & 1 & 2 & 3 & 4 & 5 & 6 & 7 & 8 & 9 \\ 
 \hline
X & 0 & 0.25 & 0.50 & 0.25 & -0.25 & 0 & -0.25 & -0.50 & 0\\ 
Y & 0 & 0.20 & 0 & -0.20 & -0.20 & 0 & 0.20 & 0 & 0 \\
Z & -2  & -2 & -2 & -2 & -2 & -2 & -2 & -2 & -2\\ 
\hline
\end{tabular}
\vspace*{0.25cm}
\caption{List of setpoints and their coordinates (in meters).}
\label{setpoints-list}
\end{table}

The experimental task is illustrated in Figure \ref{exp-task} showing the 9 setpoints considered. The disturbance generator and the camera were fixed to an aluminium arm that is fixed on the side of the pool at respectively ~60 centimeters and ~2 meters from the edge. The vehicle was set to perform station keeping at each setpoint following their numerical order.

\section{Experimental results}

\subsection{Without current disturbance}

This section presents the results of the experiments described earlier. The results shown in Tables \ref{T1-exp}-\ref{T4-exp} are associated with the experimental scenario in which no current disturbance was applied to the vehicle. We compare the proposed learning-based (LB) controller to the model-based (MB) controller using the following metrics: the root mean squared error (RMSE) on the setpoint to represent the tracking performance, the standard deviation of the RMSE to depict the smoothness of the control, the norm of the control input for power consumption, and the normalized mean return as a proxy for the sim-to-real transfer performance.

\vspace*{\fill}
\begin{table}[htbp]
\small\sf\centering
\caption{Mean RMSE without disturbance.\label{T1-exp}}
\begin{tabular}{c|cc}
\toprule
Setpoint & Model-based & Learning-based\\
\midrule
1 & 0.1189 & \textbf{0.0414}\\
2 & 0.1366 & \textbf{0.0509}\\
3 & 0.1673 & \textbf{0.0546}\\
4 & 0.1433 & \textbf{0.0544}\\
5 & 0.1085 & \textbf{0.0740}\\
6 & 0.0914 & \textbf{0.0498}\\
7 & 0.1380 & \textbf{0.0534}\\
8 & 0.1311 & \textbf{0.0463}\\
9 & 0.1106 & \textbf{0.0622}\\
\bottomrule
\end{tabular}
\end{table}

\newpage
In terms of root mean squared error (RMSE) on the setpoint (see Table \ref{T1-exp} above), the LB controller holds the smallest RMSE for every setpoint. On average, the RMSE without disturbance is $2.35$ times smaller with our LB controller.

\begin{table}[htbp]
\small\sf\centering
\caption{Std RMSE without disturbance.\label{T2-exp}}
\begin{tabular}{c|cc}
\toprule
Setpoint & Model-based & Learning-based\\
\midrule
1 & 0.0241 & \textbf{0.0147}\\
2 & 0.0325 & \textbf{0.0229}\\
3 & 0.0335 & \textbf{0.0242}\\
4 & 0.0332 & \textbf{0.0254}\\
5 & 0.0376 & \textbf{0.0295}\\
6 & 0.0247 & \textbf{0.0216}\\
7 & 0.0355 & \textbf{0.0237}\\
8 & 0.0471 & \textbf{0.0256}\\
9 & 0.0520 & \textbf{0.0285}\\
\bottomrule
\end{tabular}
\end{table}

\begin{figure}[htbp]
\centering
\includegraphics[width=\linewidth]{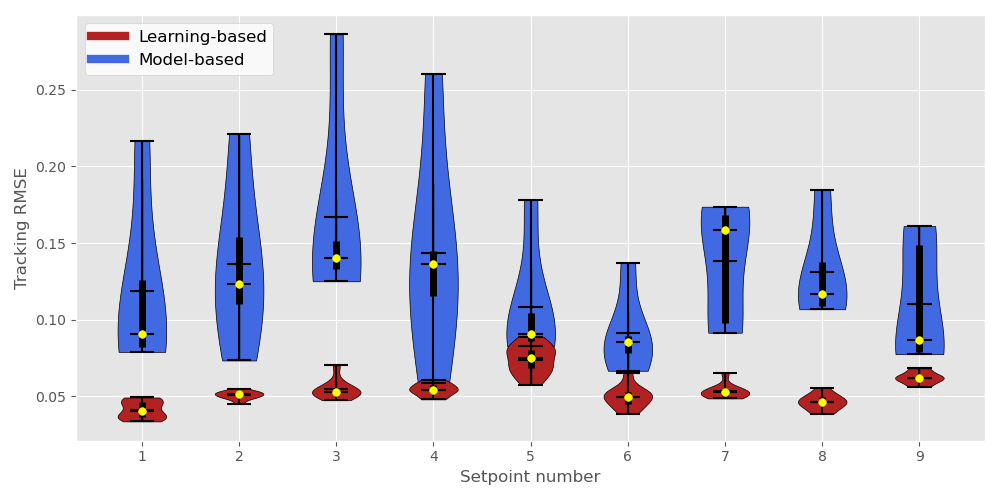}
\caption{Illustration of the experimental performance of the controllers without disturbance.}
\label{fig-perf-1-exp-1}
\end{figure}

When we take a look at the standard deviation (Std) of the RMSE (see Table \ref{T2-exp} above), which can be seen as a measure of robustness, we can also observe that the LB controller is doing better than the MB controller on every setpoint. The Std of the RMSE is again on average about 2 times smaller with our method. This tendency is furthermore perceptible in the violin plots provided in Figure \ref{fig-perf-1-exp-1} where we can see the median and quartile values computed over the 10 trials. On average, the Std of the RMSE without disturbance is $1.48$ times smaller with our LB controller.

\begin{table}[htbp]
\small\sf\centering
\caption{Normalised mean $\sum|u|$ without disturbance.\label{T3-exp}}
\begin{tabular}{c|cc}
\toprule
Setpoint & Model-based & Learning-based\\
\midrule
1 & 0.1365 & \textbf{0.1361}\\
2 & 0.1575 & \textbf{0.1507}\\
3 & 0.1806 & \textbf{0.1464}\\
4 & 0.1907 & \textbf{0.1494}\\
5 & \textbf{0.1036} & 0.1528\\
6 & \textbf{0.0880} & 0.1526\\
7 & \textbf{0.1209} & 0.1428\\
8 & \textbf{0.1141} & 0.1408\\
9 & \textbf{0.1289} & 0.1579\\
\bottomrule
\end{tabular}
\end{table}

When we take a look at the norm of the control inputs (see Table \ref{T3-exp} above), which can be seen as a measure of power consumption, we can observe a less apparent difference between the models. In fact, for the first four setpoints visited, the LB controller required smaller values of control inputs to stabilise the vehicle, while it required larger control inputs to stabilise the vehicle at the last five of the visited setpoints. On average, without disturbance, the LB controller consumed 15$\%$ more energy than the MB controller.

\begin{table}[htbp]
\small\sf\centering
\caption{Normalised mean Return without disturbance.\label{T4-exp}}
\begin{tabular}{c|cc}
\toprule
Setpoint & Model-based & Learning-based\\
\midrule
1 & 0.7698 & \textbf{0.8992}\\
2 & 0.7517 & \textbf{0.8815}\\
3 & 0.6934 & \textbf{0.8740}\\
4 & 0.7185 & \textbf{0.8738}\\
5 & 0.7871 & \textbf{0.8390}\\
6 & 0.8336 & \textbf{0.8863}\\
7 & 0.7353 & \textbf{0.8769}\\
8 & 0.7617 & \textbf{0.8929}\\
9 & 0.7785 & \textbf{0.8595}\\
\bottomrule
\end{tabular}
\end{table}

Finally, when we take a look at the mean total reward generated per episode by the agents (see Table \ref{T4-exp} above), which can be used as a metric to assess if the policy is behaving as desired, we can see that our LB controller had superior performance at every setpoint. On average, without disturbance, the gain in normalised mean return is about $15\%$ with our LB controller. In Figure \ref{fig-perf-1-exp-1}, we can see that the performance of the LB controller is better (i.e. lower error) and more robust than the MB controller (i.e. less disseminated).

\subsection{With current disturbance}

When facing current disturbances, the benefits of the proposed method are even more prominent. This is illustrated in the results provided in Tables \ref{T5-exp}-\ref{T8-exp} that are associated with the experimental scenario where a current disturbance was applied to the vehicle. Again, we can see that the LB controller presented a better performance compared to the MB controller. In terms of RMSE on the setpoint (see Table \ref{T5-exp} below), the LB controller obtained the smallest RMSE for every setpoint. On average, the mean RMSE against disturbance is $3.72$ times smaller with our LB controller.
\begin{table}[htbp]
\small\sf\centering
\caption{Mean RMSE with disturbance.\label{T5-exp}}
\begin{tabular}{c|cc}
\toprule
Setpoint & Model-based & Learning-based\\
\midrule
1 & 0.3723 & \textbf{0.0882}\\
2 & 0.3587 & \textbf{0.0844}\\
3 & 0.3420 & \textbf{0.0731}\\
4 & 0.3645 & \textbf{0.0783}\\
5 & 0.3686 & \textbf{0.1120}\\
6 & 0.3647 & \textbf{0.0797}\\
7 & 0.3648 & \textbf{0.1149}\\
8 & 0.3606 & \textbf{0.1385}\\
9 & 0.3826 & \textbf{0.1100}\\
\bottomrule
\end{tabular}
\end{table}

In terms of the Std of the RMSE (see Table \ref{T6-exp} below), the LB controller outperformed the MB controller at every setpoint. On average, the Std of the RMSE against disturbance is $2.96$ times smaller with our LB controller.

\begin{table}[htbp]
\small\sf\centering
\caption{Std RMSE with disturbance.\label{T6-exp}}
\begin{tabular}{c|cc}
\toprule
Setpoint & Model-based & Learning-based\\
\midrule
1 & 0.1573 & \textbf{0.0289}\\
2 & 0.0947 & \textbf{0.0330}\\
3 & 0.0888 & \textbf{0.0285}\\
4 & 0.1068 & \textbf{0.0294}\\
5 & 0.0893 & \textbf{0.0451}\\
6 & 0.1005 & \textbf{0.0277}\\
7 & 0.0969 & \textbf{0.0319}\\
8 & 0.0987 & \textbf{0.0530}\\
9 & 0.1206 & \textbf{0.0444}\\
\bottomrule
\end{tabular}
\end{table}

When considering the norm of the control inputs (see Table \ref{T7-exp}), we can now see that the LB controller required larger control inputs at all setpoints to stabilise the vehicle compared to the first environmental conditions. We believe that this result is explained by the successful detection of variations in the current disturbance. In fact, the attenuation of low-frequency disturbance is inversely proportional to the integral gain. Maximizing the integral gain is a good heuristic to obtain a PID controller with good disturbance rejection. The LB controller is able to detect this change and increase the control parameters resulting in higher control inputs. Nevertheless, given the adaptive pole-placement design \citep{Chaffre2021DirectAP}, the resulting gains of the PID controller are positively correlated. Thus, the derivative gain will also increase, which decreases stability margins. However, for pole values lower than 1, the proportional and integral gains vary exponentially while the derivative gain varies linearly \citep{Chaffre2021DirectAP}. The LB controller successfully increases the proportional and integral gains while maintaining the same order derivative gain. This results in better disturbance rejection with similar smoothness in the control of the vehicle as suggested by the lower RMSE and std RMSE.
On average, the LB controller consumed 9$\%$ more energy than the MB controller.
\begin{table}[htbp]
\small\sf\centering
\caption{Normalised mean $\sum|u|$ with disturbance.\label{T7-exp}}
\begin{tabular}{c|cc}
\toprule
Setpoint & Model-based & Learning-based\\
\midrule
1 & \textbf{0.1449} & 0.1816\\
2 & \textbf{0.1407} & 0.1739\\
3 & \textbf{0.1362} & 0.1407\\
4 & \textbf{0.1406} & 0.1475\\
5 & \textbf{0.1424} & 0.1637\\
6 & \textbf{0.1417} & 0.1311\\
7 & \textbf{0.1404} & 0.1389\\
8 & \textbf{0.1424} & 0.1874\\
9 & \textbf{0.1530} & 0.1717\\
\bottomrule
\end{tabular}
\end{table}

Finally, when taking into account the mean total reward generated per episode by the agents (see Table \ref{T8-exp}), we can that the LB controller also outperforms the MB controller at every setpoint. The normalised mean return of the LB controller was on average about $1.36$ times higher than the MB controller. It is worth observing that the MB controller was not able to stabilise the vehicle while the LB controller was successfully completing the task.
\begin{table}[htbp]
\small\sf\centering
\caption{Normalised mean Return with disturbance.\label{T8-exp}}
\begin{tabular}{c|cc}
\toprule
Setpoint & Model-based & Learning-based\\
\midrule
1 & 0.5260 & \textbf{0.7972}\\
2 & 0.6642 & \textbf{0.8008}\\
3 & 0.6648 & \textbf{0.8384}\\
4 & 0.6309 & \textbf{0.8348}\\
5 & 0.5326 & \textbf{0.7753}\\
6 & 0.5646 & \textbf{0.8309}\\
7 & 0.5461 & \textbf{0.7752}\\
8 & 0.6478 & \textbf{0.7486}\\
9 & 0.4831 & \textbf{0.7840}\\
\bottomrule
\end{tabular}
\end{table}

\begin{figure}[htbp]
\centering
\includegraphics[width=\linewidth]{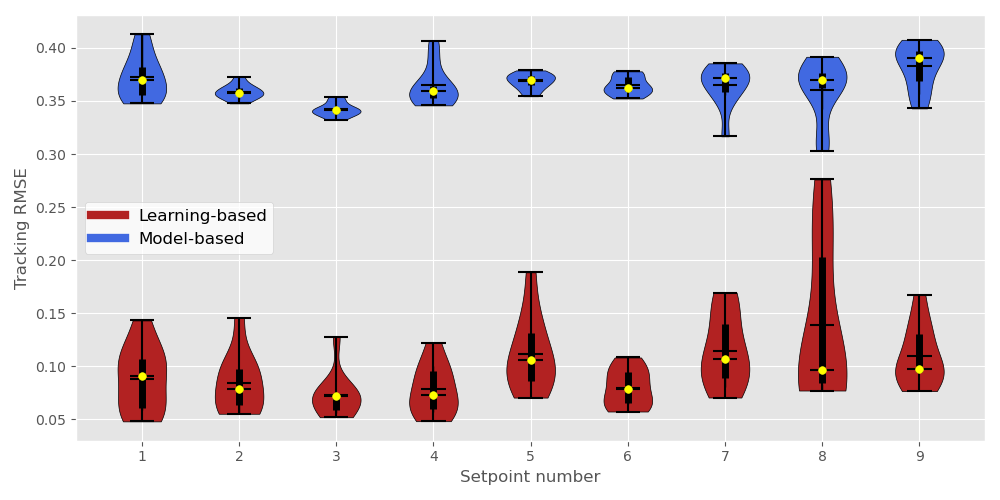}
\caption{Illustration of the experimental performance of the controllers with current disturbance.}
\label{fig-perf-2-exp-2}
\end{figure}

\newpage
This tendency is shown in the violin plots provided in Figure \ref{fig-perf-2-exp-2} where we can again see the median and quartile values computed over the 10 trials. The difference in performance is furthermore important against current disturbance.

\begin{figure}[htbp]
\centering
\includegraphics[width=\linewidth]{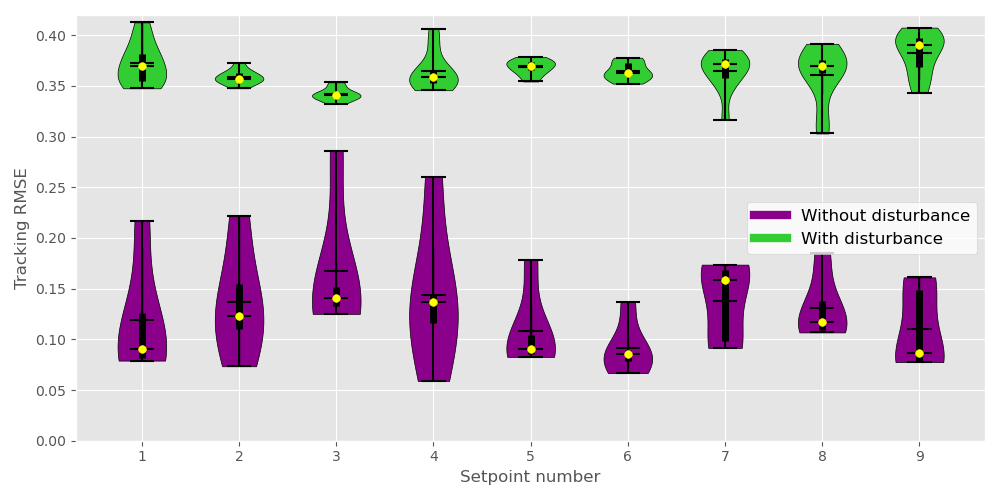}
\caption{Example of the experimental performance of the MB controller without and with current disturbance.}
\label{fig-perf-3}
\end{figure}

\begin{figure}[htbp]
\centering
\includegraphics[width=\linewidth]{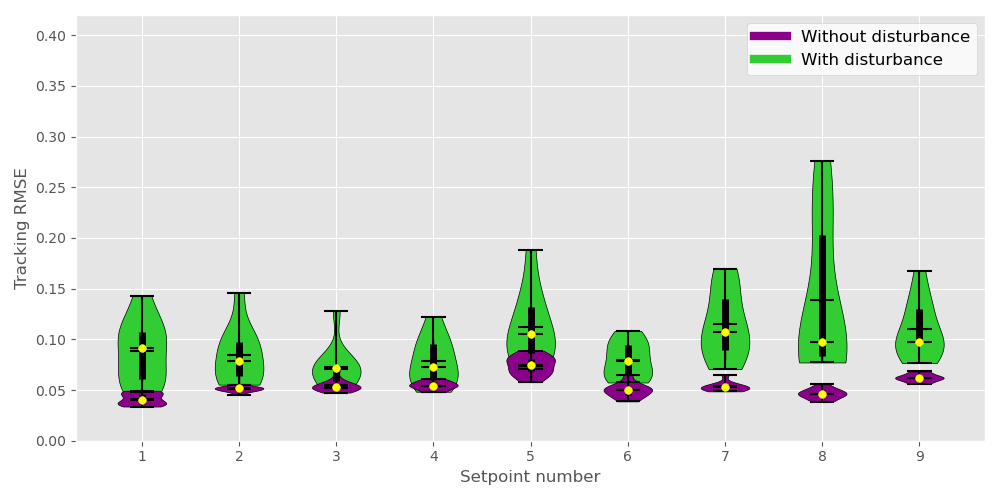}
\caption{Example of the experimental performance of the LB controller without and with current disturbance.}
\label{fig-perf-4}
\end{figure}

Figures \ref{fig-perf-3} and \ref{fig-perf-4} below provide the opportunity to compare the violin plots of the results for each controller. We can see that despite using a sub-optimal simulated model of the AUV, the learning-based policy performed notably better when transferred to the real platform compared to its nonadaptive optimal counterpart.

\begin{figure}[htbp]
\centering
\begin{subfigure}{0.5\textwidth}
    \centering
    \includegraphics[width=0.75\textwidth]{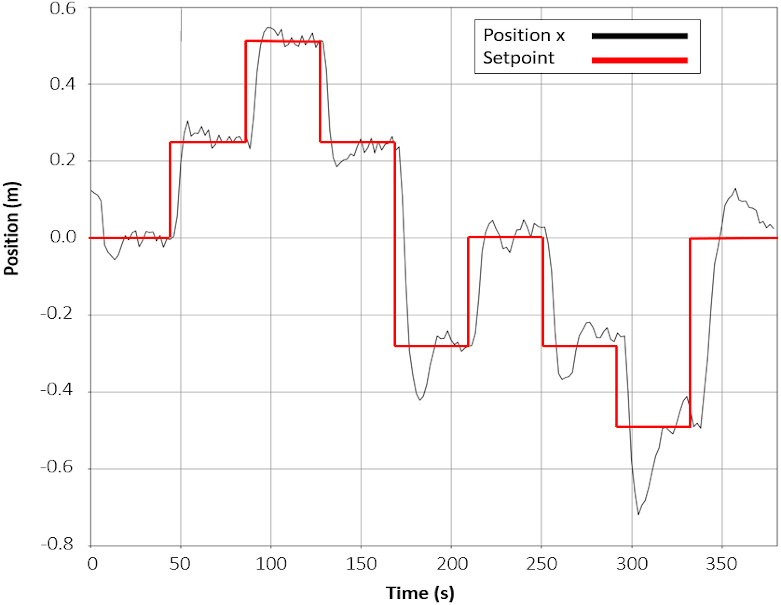}
    \captionsetup{justification=centering}
    \caption{MB controller.}
    \label{mb-x}
\end{subfigure}
\begin{subfigure}{0.5\textwidth}
    \centering
    \includegraphics[width=0.75\textwidth]{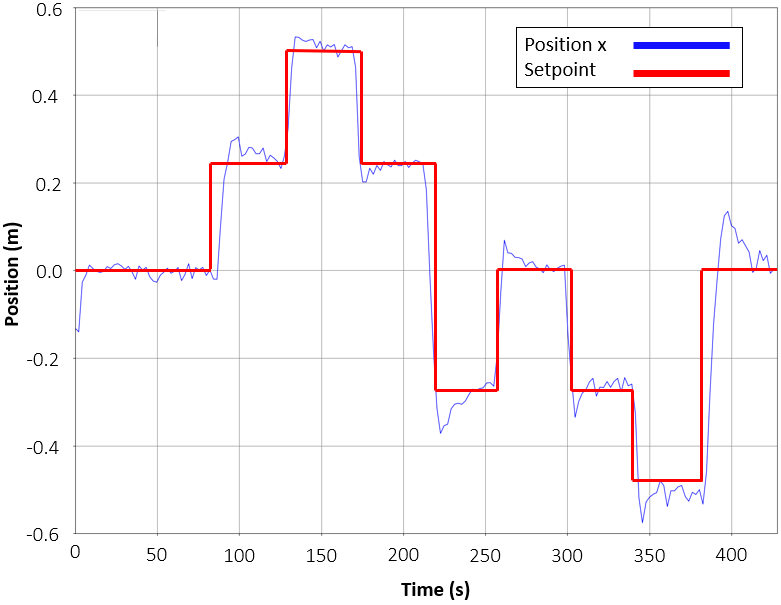}
    \captionsetup{justification=centering}
    \caption{LB controller.}
    \label{lb-x}
\end{subfigure}
\captionsetup{justification=centering}
\caption{Evolution of the position $X$.}
\label{traj1}
\end{figure}
\begin{figure}[htbp]
\centering
\begin{subfigure}{0.5\textwidth}
    \centering
    \includegraphics[width=0.75\textwidth]{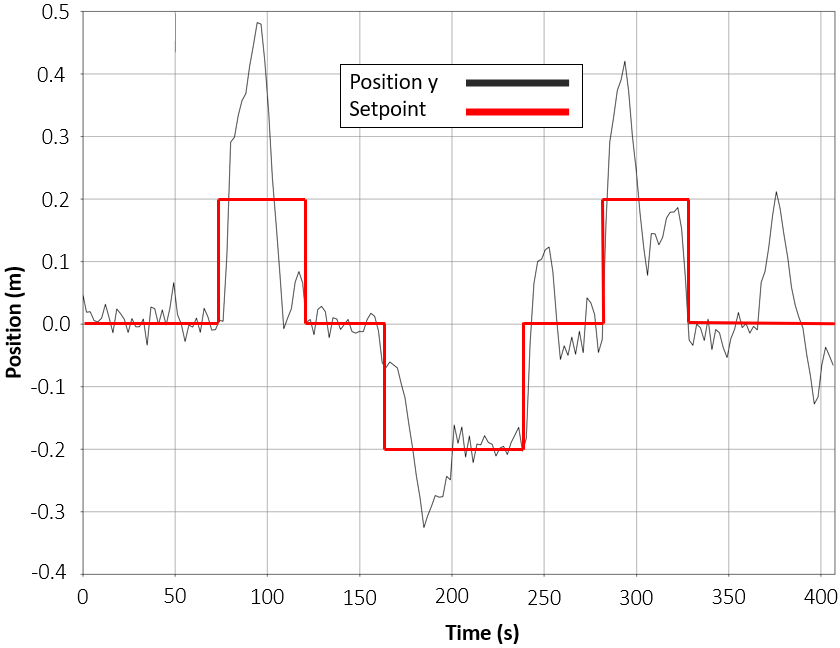}
    \captionsetup{justification=centering}
    \caption{MB controller.}
    \label{mb-y}
\end{subfigure}
\begin{subfigure}{0.5\textwidth}
    \centering
    \includegraphics[width=0.75\textwidth]{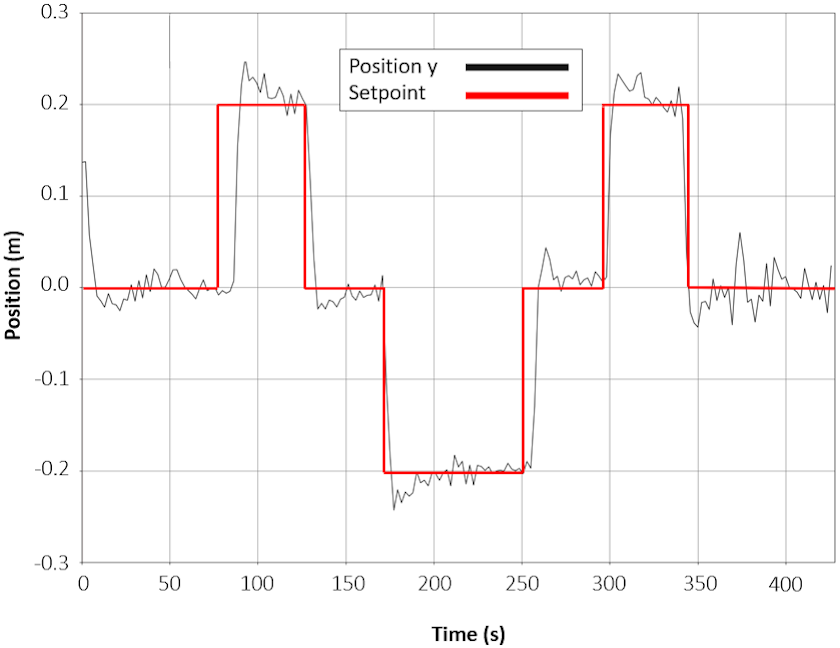}
    \captionsetup{justification=centering}
    \caption{LB controller.}
    \label{lb-y}
\end{subfigure}
\captionsetup{justification=centering}
\caption{Evolution of the position $Y$.}
\label{traj2}
\end{figure}
\begin{figure}[htbp]
\centering
\begin{subfigure}{0.5\textwidth}
    \centering
    \includegraphics[width=0.75\textwidth]{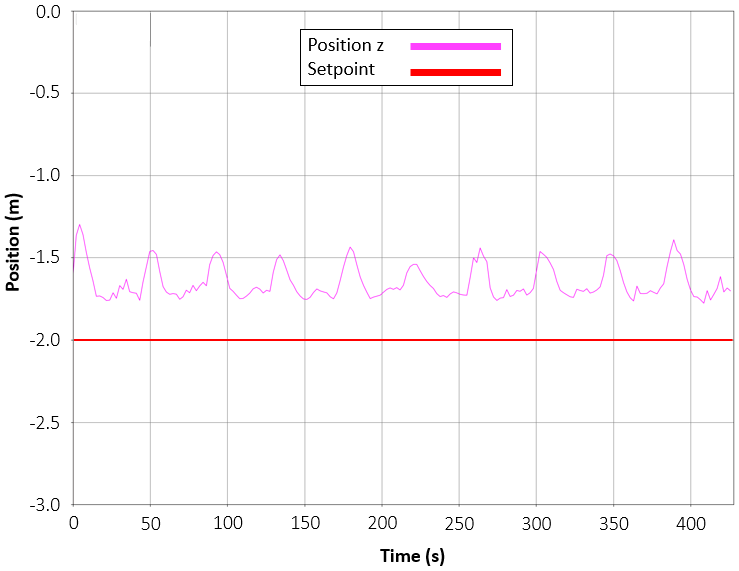}
    \captionsetup{justification=centering}
    \caption{MB controller.}
    \label{mb-z}
\end{subfigure}
\begin{subfigure}{0.5\textwidth}
    \centering
    \includegraphics[width=0.75\textwidth]{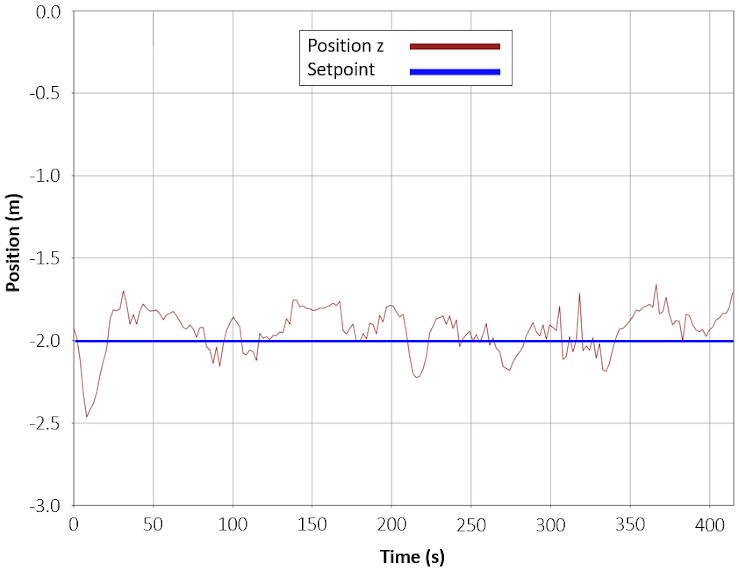}
    \captionsetup{justification=centering}
    \caption{LB controller.}
    \label{lb-z}
\end{subfigure}
\captionsetup{justification=centering}
\caption{Evolution of the position $Z$.}
\label{traj3}
\end{figure}
\begin{figure}[htbp]
\centering
\begin{subfigure}{0.5\textwidth}
    \centering
    \includegraphics[width=0.75\textwidth]{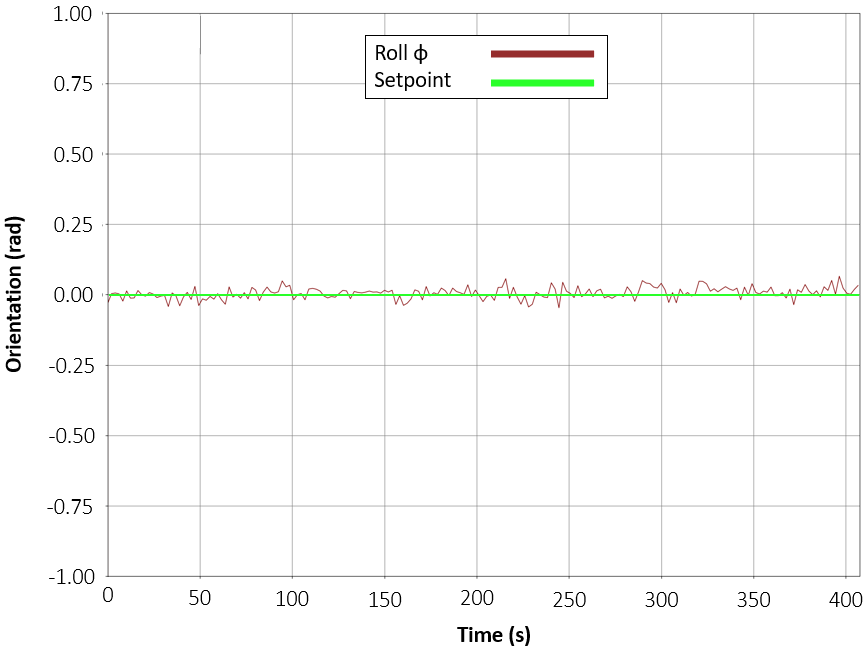}
    \captionsetup{justification=centering}
    \caption{MB controller.}
    \label{mb-roll}
\end{subfigure}
\begin{subfigure}{0.5\textwidth}
    \centering
    \includegraphics[width=0.75\textwidth]{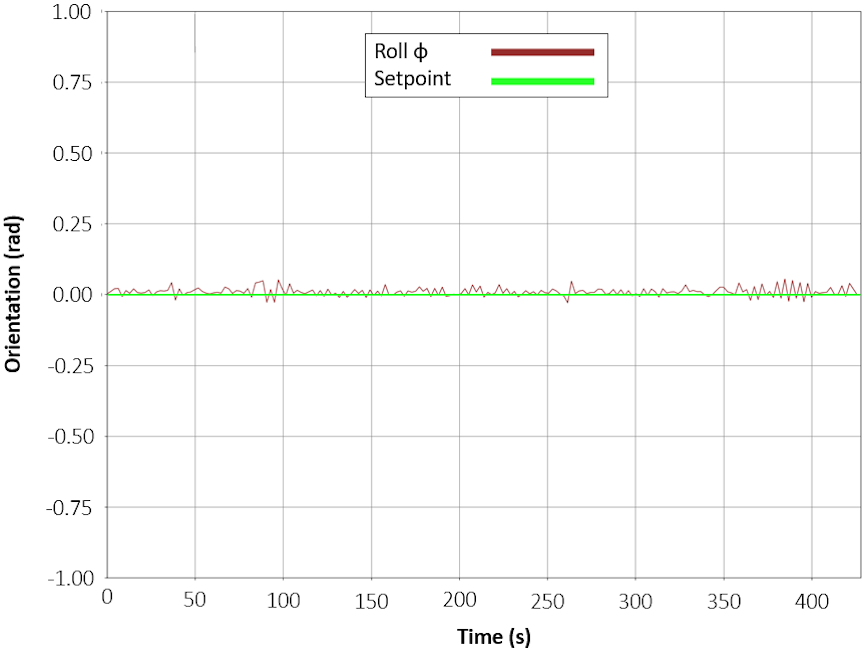}
    \captionsetup{justification=centering}
    \caption{LB controller.}
    \label{lb-roll}
\end{subfigure}
\captionsetup{justification=centering}
\caption{Evolution of the roll $\psi$.}
\label{traj4}
\end{figure}
\begin{figure}[htbp]
\centering
\begin{subfigure}{0.5\textwidth}
    \centering
    \includegraphics[width=0.75\textwidth]{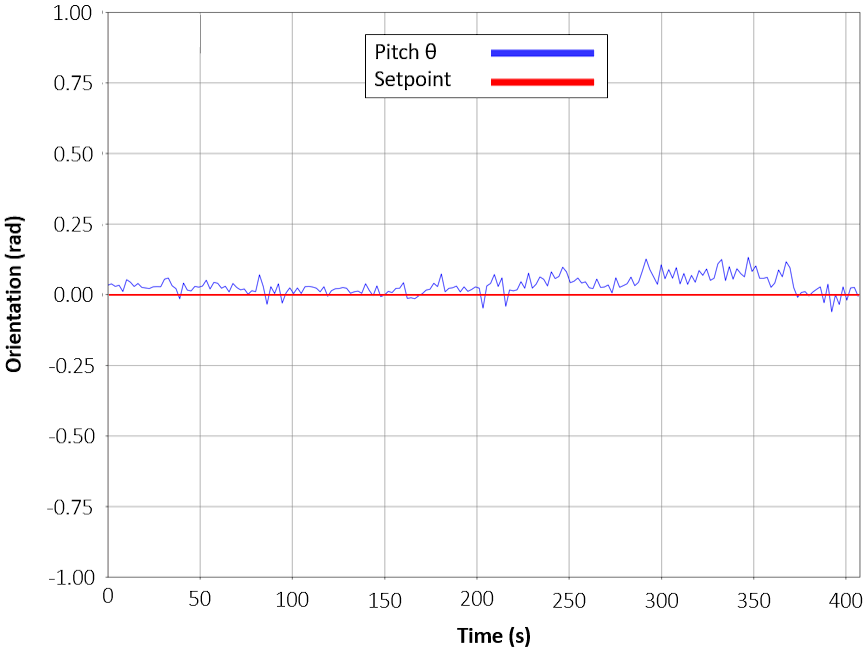}
    \captionsetup{justification=centering}
    \caption{MB controller.}
    \label{mb-pitch}
\end{subfigure}
\begin{subfigure}{0.5\textwidth}
    \centering
    \includegraphics[width=0.75\textwidth]{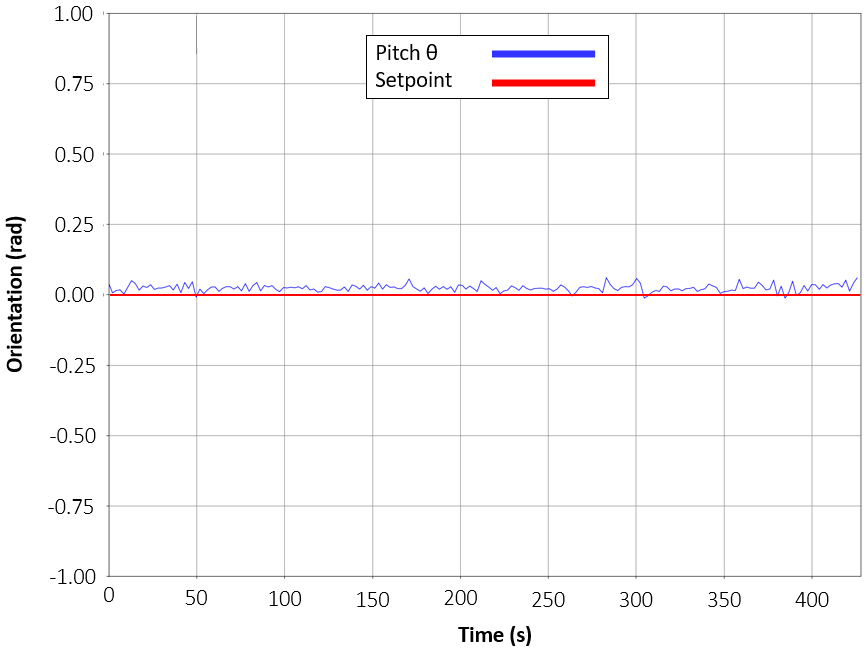}
    \captionsetup{justification=centering}
    \caption{LB controller.}
    \label{lb-pitch}
\end{subfigure}
\captionsetup{justification=centering}
\caption{Evolution of the pitch $\theta$.} \label{traj5}
\end{figure}
\begin{figure}[htbp]
\centering
\begin{subfigure}{0.5\textwidth}
    \centering
    \includegraphics[width=0.75\textwidth]{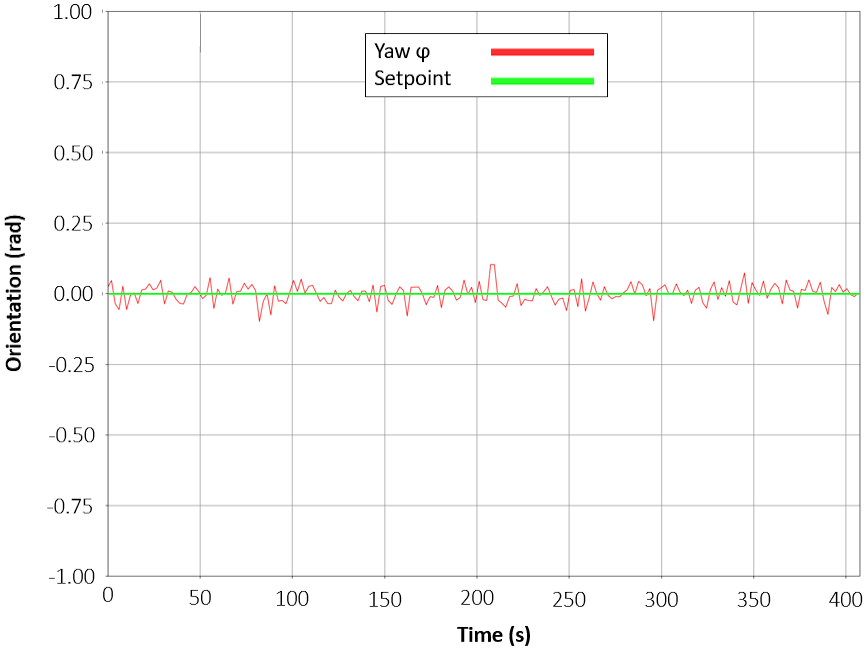}
    \captionsetup{justification=centering}
    \caption{MB controller.}
    \label{mb-yaw}
\end{subfigure}
\begin{subfigure}{0.5\textwidth}
    \centering
    \includegraphics[width=0.75\textwidth]{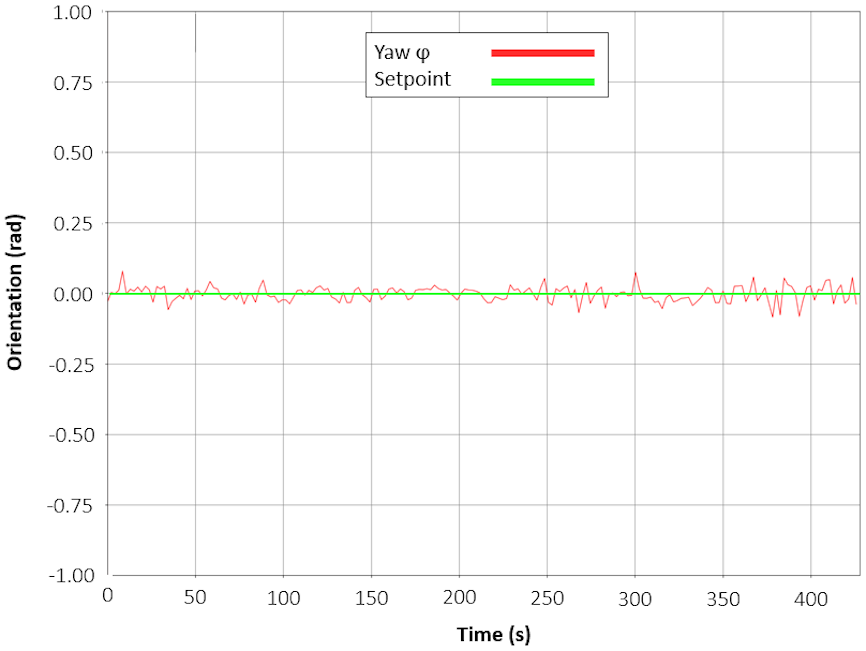}
    \captionsetup{justification=centering}
    \caption{LB controller.}
    \label{lb-yaw}
\end{subfigure}
\captionsetup{justification=centering}
\caption{Evolution of the yaw $\phi$.}
\label{traj6}
\end{figure}

Figures \ref{traj1}-\ref{traj6} show the trajectories performed by both controllers for each DoF during an episode with current disturbance. \textcolor{black}{These trajectories might not be representative of the mean performance of the controllers outlined in the previous tables, but they were chosen as they provide great insights into the controllers' behaviour.} Overall, we are able to observe that the proposed learning-based adaptive controller displayed a lower overshoot and is better at tracking the desired 
trajectory.

In Figures \ref{traj1}-\ref{traj3}, we can see that the overshoot on the setpoint is smaller with the LB controller and, in this particular example, the position in $Z$ is not being regulated by the MB controller.

In Figures \ref{traj4}-\ref{traj6}, we can see that in terms of Euler angles, the LB controller also displays better trajectories. \textcolor{black}{To conclude, we have presented here the results of an experimental evaluation. We evaluated the two controllers on a multi-station keeping task and in two distinct scenarios: without and with current disturbance. We have presented the resulting outcomes of this evaluation as the mean values of multiple key performance indicators obtained over 10 trials for each controller, emphasising about 280 minutes of real-life operating time. We have shown experimentally that the proposed LB adaptive controller consistently outperformed the MB optimal controller.}

\newpage

\vspace*{\fill}

\newpage
\section{Discussion}

Learning-based adaptive control provides an efficient way to cope with process variations by providing a model-free adjustment mechanism. However, so far their success at solving difficult AUV processes has been limited, mostly due to the partial-observability of underwater environments. We have argued that the key to a successful sim-to-real transfer is to obtain good estimates of the Q-Value function via Domain Randomisation \citep{chen2022understanding} and Maximum Entropy DRL \citep{DBLP:conf/iclr/EysenbachL22}.

We have provided a methodology to design a learning-based adaptive control system on the basis of the PID control law, which represents the vast majority of in-use AUV control systems. We described how to combine this control structure with the Soft Actor-Critic with Automatic Entropy Adjustment algorithm that optimises a value and policy function, both represented by DNNs. Combining model-based control and model-free learning, we are able to compensate for unobservable current disturbances.

Our main experimental validation was in the domain of manoeuvring tasks for AUV. 
Despite being trained on a different model of the vehicle under simulations, the resulting policy was still able to regulate the vehicle and displayed performance between 2 and 3 times higher (in terms of setpoint regulation) compared to the nonadaptive optimal model-based controller. This was possible thanks to the proposed learning-based architecture where DRL is used to learn to adapt to the overall dynamics of the process rather than learning the dynamics of a specific vehicle/process (i.e. end-to-end DRL). This approach grants the control system the ability to learn how to adjust the control parameters with respect to changes in the error signals, making it easier to transfer to a slightly different vehicle/process.

One question that deserves future investigation is the relationship between the process observability and the distribution shift problem in RL \citep{Ghosh2021WhyGI} \citep{li2023efficient}. If this relationship were known, we could greatly reduce the overestimation problem \citep{Kumar2019StabilizingOQ} of Deep Q-Learning. Some candidates for that are Distribution Constraints via Lyapunov Theory \citep{pmlr-v162-kang22a}, improved Regularisation \citep{eysenbach2022a}, \citep{52387}, or Implicit Q-Learning \citep{kostrikov2022offline}.

\section{Conclusions}

This paper investigated the application of learning-based adaptive control in the context of AUV disturbance rejection, yielding several noteworthy contributions, summarised as:

\begin{itemize}
\item a novel learning-based adaptive control architecture was introduced, designed for utilisation alongside traditional feedback control methods, such as PID controllers, resulting in a controller that is adaptable to changes at the same time that it maintains a backbone grounded on physical modelling of the plant.

\item a comprehensive empirical evaluation was conducted by implementing and assessing the proposed learning-based adaptive controller alongside its nonadaptive, model-based counterpart on an actual AUV platform. Remarkably, despite sharing an identical controller structure, the learning-based approach exhibited substantial performance enhancements.

\item this research contributed with an analysis of the transferability of the policies learned from simulation to the physical plant, wherein the learning-based adaptive controller, initially trained on a dissimilar vehicle model, demonstrated the capability to effectively stabilise the AUV in a real-world context, underscoring its adaptability and generalisability.

\item An exploration into the correlation between the complexity levels of source and target domains led to the identification of a pivotal factor: domain randomisation. We observed that randomising environmental complexity, quantified by factors such as sea current disturbance amplitude and task difficulty, mitigated policy variance, thus elucidating a key mechanism contributing to the improved sim-to-real transfer.
\end{itemize}

Additionally, to facilitate the transition from simulation to practical deployment, we introduced a refinement to the SAC algorithm, incorporating automatic temperature parameter adjustment. This innovation obviates the need for intensive and empirical reward scale parameter tuning, enhancing the method's usability and efficiency.

Future work shall concentrate on evaluating the proposed methods in an industrial-level application of a AUV operating in an open sea environment. Not only will this provide stronger evidence for the value of the proposed work, but also it will allow the possibility to incorporate nonlinear model-based control structures so as to cope with underactuated situations in a more challenging environment.

\newpage
\section*{Acknowledgement}
The authors would like to thank Dr. Estelle Chauveau from CEMIS, the Naval Group Research’s Centre of Excellence for Information, Human Factors and Signature Management, for helpful discussions and technical advice. This work was supported in part by SENI, the research laboratory between Naval Group and ENSTA Bretagne.

\bibliographystyle{SageH}
\bibliography{bibliography}

\end{document}